\documentclass[10pt,twocolumn,letterpaper]{article}

\usepackage{cvpr}              

\usepackage{graphicx}
\usepackage{amsmath}
\usepackage{amssymb}
\usepackage{amsthm}
\usepackage{booktabs}
\usepackage{adjustbox}
\usepackage{multicol}
\usepackage{algpseudocode}

\usepackage[pagebackref,breaklinks,colorlinks]{hyperref}

\usepackage[capitalize]{cleveref}
\crefname{section}{Sec.}{Secs.}
\Crefname{section}{Section}{Sections}
\Crefname{table}{Table}{Tables}
\crefname{table}{Tab.}{Tabs.}

\usepackage{caption}

\def\cvprPaperID{76} 
\def\confName{CVPR}
\def\confYear{2023}

\usepackage[vlined,ruled,linesnumbered]{algorithm2e}
\usepackage{graphics} 
\usepackage{rotating}
\usepackage{color}
\usepackage{enumerate}
\usepackage[T1]{fontenc}
\usepackage{multirow}
\usepackage{amsfonts}
\usepackage{amsmath}
\usepackage{amssymb}
\usepackage{xstring}
\usepackage{xcolor}
\usepackage{prettyref}
\usepackage{enumitem}
\usepackage{xspace}
\usepackage{bm}
\graphicspath{{./figures/}}
\newrefformat{prob}{Problem\,\ref{#1}}
\newrefformat{def}{Definition\,\ref{#1}}
\newrefformat{sec}{Section\,\ref{#1}}
\newrefformat{sub}{Section\,\ref{#1}}
\newrefformat{prop}{Proposition\,\ref{#1}}
\newrefformat{app}{Appendix\,\ref{#1}}
\newrefformat{alg}{Algorithm\,\ref{#1}}
\newrefformat{cor}{Corollary\,\ref{#1}}
\newrefformat{thm}{Theorem\,\ref{#1}}
\newrefformat{lem}{Lemma\,\ref{#1}}
\newrefformat{fig}{Fig.\,\ref{#1}}
\newrefformat{tab}{Table\,\ref{#1}}

\newtheorem{theorem}{Theorem}

\newtheorem{proposition}[theorem]{Proposition}
\newtheorem{remark}[theorem]{Remark}

\newcommand{\bdmath}{\begin{dmath}}
\newcommand{\edmath}{\end{dmath}}
\newcommand{\beq}{\begin{equation}}
\newcommand{\eeq}{\end{equation}}
\newcommand{\bdm}{\begin{displaymath}}
\newcommand{\edm}{\end{displaymath}}
\newcommand{\bea}{\begin{eqnarray}}
\newcommand{\eea}{\end{eqnarray}}
\newcommand{\beal}{\beq \begin{array}{ll}}
\newcommand{\eeal}{\end{array} \eeq}
\newcommand{\beas}{\begin{eqnarray*}}
\newcommand{\eeas}{\end{eqnarray*}}
\newcommand{\ba}{\begin{array}}
\newcommand{\ea}{\end{array}}
\newcommand{\bit}{\begin{itemize}}
\newcommand{\eit}{\end{itemize}}
\newcommand{\ben}{\begin{enumerate}}
\newcommand{\een}{\end{enumerate}}

\newcommand{\calX}{{\cal X}}
\newcommand{\calY}{{\cal Y}}
\newcommand{\calZ}{{\cal Z}}

\renewcommand{\boldsymbol}[1]{{\bm #1}}

\newcommand{\hide}[1]{}

\newcommand{\hiddenText}{{\color{gray} hidden text.}}
\newcommand{\hideWithText}[1]{\hiddenText}

\newcommand{\kron}{\otimes}

\newcommand{\subject}{\text{ subject to }}

\newcommand{\norm}[1]{\left\| #1 \right\|}

\newcommand{\tran}{^{\mathsf{T}}}

\newcommand{\inv}{^{-1}}

\newcommand{\eye}{{\mathbf I}}

\newcommand{\Real}[1]{ { {\mathbb R}^{#1} } }

\newcommand{\SEthree}{\ensuremath{\mathrm{SE}(3)}\xspace}

\newcommand{\SOthree}{\ensuremath{\mathrm{SO}(3)}\xspace}

\newcommand{\vf}{\boldsymbol{f}}

\newcommand{\vv}{\boldsymbol{v}}

\newcommand{\vy}{\boldsymbol{y}}

\newcommand{\scenario}[1]{{\smaller \sf#1}\xspace}

\newcommand{\blue}[1]{{\color{blue}#1}}

\newcommand{\linkToPdf}[1]{\href{#1}{\blue{(pdf)}}}
\newcommand{\linkToPpt}[1]{\href{#1}{\blue{(ppt)}}}
\newcommand{\linkToCode}[1]{\href{#1}{\blue{(code)}}}
\newcommand{\linkToWeb}[1]{\href{#1}{\blue{(web)}}}
\newcommand{\linkToVideo}[1]{\href{#1}{\blue{(video)}}}
\newcommand{\linkToMedia}[1]{\href{#1}{\blue{(media)}}}
\newcommand{\award}[1]{\xspace} 

\renewcommand{\norm}[1]{\Vert #1 \Vert}

\newcommand{\vectorize}[1]{\mathrm{vec}\parentheses{#1}}
\newcommand{\Fnorm}[1]{\Vert #1 \Vert_{\mathrm{F}}}

\newcommand{\cbrace}[1]{\left\{#1\right\}}
\newcommand{\sym}[1]{\mathbb{S}^{#1}}

\newcommand{\barq}{\bar{q}}

\newcommand{\bmat}{\left[ \begin{array}}
\newcommand{\emat}{\end{array}\right]}

\newcommand{\parentheses}[1]{\left(#1\right)}

\newcommand{\abs}[1]{\left|#1\right|}

\newcommand{\haty}{\hat{y}}

\newcommand{\probof}[1]{\mathbb{P}\left[#1\right]}
\newcommand{\Feps}{F^{\epsilon}}
\newcommand{\Fepsball}{\Feps_{\mathrm{ball}}}
\newcommand{\Fepsellipse}{\Feps_{\mathrm{ellipse}}}
\newcommand{\tcalY}{\tilde{\calY}}
\newcommand{\floor}[1]{\lfloor #1 \rfloor}

\newcommand{\lmo}{\scenario{LM-O}}

\newcommand{\gtball}{\scenario{gt-ball}}
\newcommand{\gtellipse}{\scenario{gt-ellipse}}
\newcommand{\frcnnball}{\scenario{frcnn-ball}}
\newcommand{\frcnnellipse}{\scenario{frcnn-ellipse}}
\newcommand{\purse}{\scenario{PURSE}}
\newcommand{\ransag}{\scenario{RANSAG}}
\newcommand{\pthreep}{\scenario{P3P}}
\newcommand{\phipeak}{\phi_{\mathrm{peak}}}
\newcommand{\phicov}{\phi_{\mathrm{cov}}}
\newcommand{\pnp}{\scenario{PnP}}
\newcommand{\Seps}{S^{\epsilon}}
\newcommand{\ransac}{\scenario{RANSAC}}
\newcommand{\Rgt}{R_{\mathrm{gt}}}
\newcommand{\tgt}{t_{\mathrm{gt}}}
\newcommand{\sgt}{s_{\mathrm{gt}}}

\begin{document}


\title{\vspace{-18mm} Object Pose Estimation with Statistical Guarantees: \\ Conformal Keypoint Detection and Geometric Uncertainty Propagation \vspace{-6mm}}

\author{Heng Yang and Marco Pavone\\
NVIDIA Research
}

\twocolumn[{%
\renewcommand\twocolumn[1][]{#1}%
\maketitle
\vspace{-8mm}
\begin{minipage}{\textwidth}
\includegraphics[width=\textwidth]{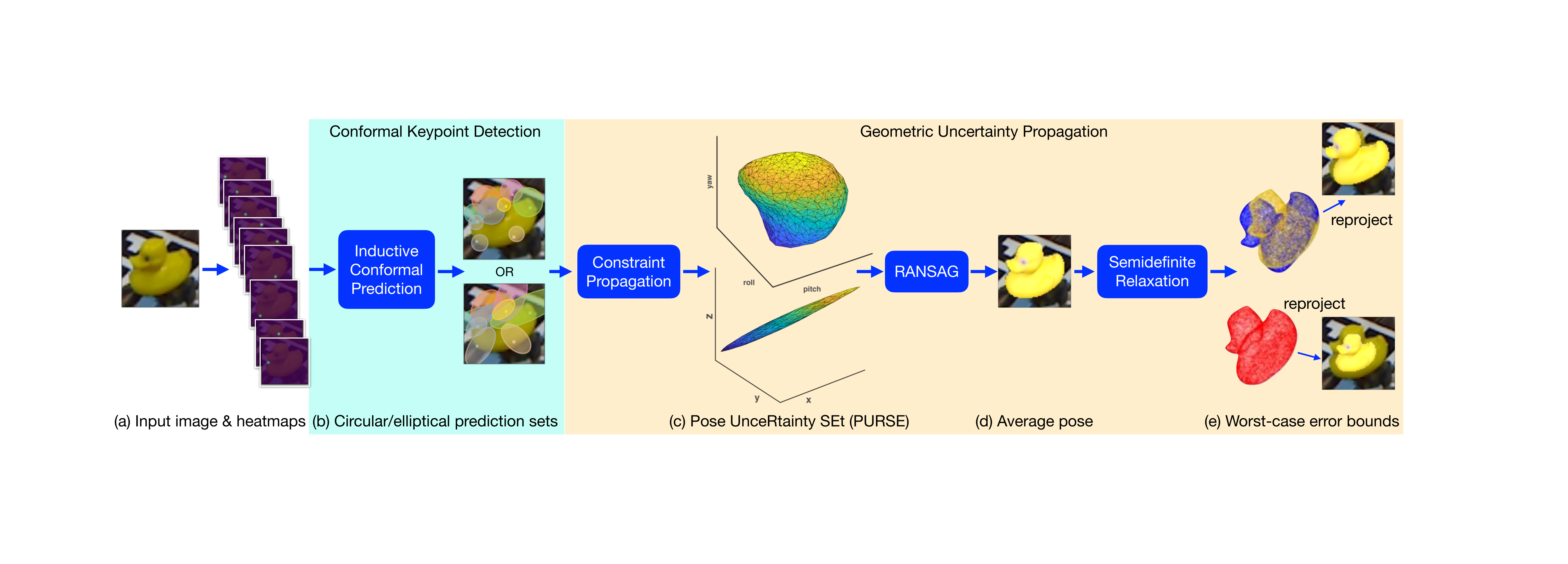}
\captionof{figure}{Probabilistically correct object pose estimation. Given (a) an input image and heatmap detections of the object semantic keypoints, our method first \emph{conformalizes} the heatmaps into (b) circular or elliptical prediction sets that \emph{guarantee} probabilistic coverage of the groundtruth keypoints (\eg, $90\%$). Our method then propagates the uncertainty in the keypoints to the object pose, forming (c) a Pose UnceRtainty SEt (\purse) that contains the groundtruth pose with the same probability. 
We develop RANdom SAmple averaGing (\ransag) to sample from {\purse} and generate (d) an average pose, and apply semidefinite relaxation to compute (e) worst-case error bounds: the blue duck attains the worst rotation error \wrt the (average pose) yellow duck; the red duck attains the worst translation error. Code available: \url{https://github.com/NVlabs/ConformalKeypoint}.\label{fig:methodoverview}}
\vspace{2mm}
\end{minipage}


}]


\begin{abstract}
\vspace{-4mm}
The two-stage object pose estimation paradigm first detects semantic keypoints on the image and then estimates the 6D pose by minimizing reprojection errors. 
Despite performing well on standard benchmarks, existing techniques offer no {provable guarantees} on the quality and uncertainty of the estimation. 
In this paper, we inject two fundamental changes, namely \emph{conformal keypoint detection} and \emph{geometric uncertainty propagation}, into the two-stage paradigm and propose the first pose estimator that endows an estimation with \emph{provable and computable worst-case error bounds}.
On one hand, conformal keypoint detection applies the statistical machinery of \emph{inductive conformal prediction} to convert 
heuristic keypoint detections into circular or elliptical prediction sets that cover the groundtruth keypoints with a user-specified marginal probability (\eg, $90\%$). Geometric uncertainty propagation, on the other, propagates the geometric constraints on the keypoints to the 6D object pose, leading to a \emph{Pose UnceRtainty SEt (\purse)} that guarantees coverage of the groundtruth pose with the same probability. The {\purse}, however, is a nonconvex set that does not directly lead to estimated poses and uncertainties. Therefore, we develop \emph{RANdom SAmple averaGing (\ransag)} to compute an {average} pose and apply semidefinite relaxation to upper bound the worst-case errors between the average pose and the groundtruth. On the LineMOD Occlusion dataset we demonstrate: (i) the \purse covers the groundtruth with valid probabilities; (ii) the worst-case error bounds provide correct uncertainty quantification; and (iii) the average pose achieves better or similar accuracy as representative methods based on sparse keypoints.
\end{abstract}
\vspace{-6mm}
\section{Introduction}
Estimating object poses from images is a fundamental problem in computer vision and finds extensive applications in augmented reality~\cite{klein07mar-ptam}, autonomous driving~\cite{shi21rss-optimal}, robotic manipulation~\cite{manuelli19isrr-kpam}, and space robotics~\cite{chen19iccvw-satellite}. One of the most popular paradigms for object pose estimation is a \emph{two-stage} pipeline~\cite{peng19cvpr-pvnet,pavlakos17icra-semantic,schmeckpeper22jfr-semantic,tekin18cvpr-yolo,zakharov19iccv-dpod,sun22cvpr-onepose,shi22arxiv-optimal,chen20cvpr-backproppnp}, where the first stage detects (semantic) {keypoints} of the objects on the image, and the second stage computes the object pose by solving an optimization known as \emph{Perspective-$n$-Points} (\pnp) that minimizes {reprojection errors} of the detected keypoints.

\emph{Safety-critical} applications call for \emph{provably correct} computer vision algorithms. Existing algorithms in the two-stage paradigm (reviewed in Section~\ref{sec:related-work}), however, provide few performance guarantees on the quality of the estimated poses, due to three challenges. (C1) It is difficult to ensure the detected keypoints (typically from neural networks) are close to the groundtruth keypoints. In practice, the first stage often outputs keypoints that are arbitrarily wrong, known as \emph{outliers}. (C2) Robust estimation is employed in the second stage to reject outliers, leading to nonconvex optimizations. Fast heuristics such as \ransac~\cite{fischler81acm-ransac} are widely adopted to find an approximate solution but they cannot guarantee global optimality and often fail without notice. (C3) There is no provably correct \emph{uncertainty quantification} of the estimation, notably, a \emph{formal worst-case error bound} between the estimation and the groundtruth. Though recent work~\cite{yang22pami-certifiably} proposed convex relaxations to certify global optimality of~\ransac and addressed (C2), it cannot ensure correct estimation as the optimal pose may be far away from the correct pose when the keypoints are unreliable.

{\bf Contributions}. We propose a two-stage object pose estimation framework with \emph{statistical guarantees}, illustrated in Fig.~\ref{fig:methodoverview}. Given an input image, we assume a neural network~\cite{pavlakos17icra-semantic} is available to generate \emph{heatmap} predictions of the object keypoints (Fig.~\ref{fig:methodoverview}(a)). Our framework then proceeds in two stages, namely \emph{conformal keypoint detection} (Section~\ref{sec:keypoint:conformal}) and \emph{geometric uncertainty propagation} (Section~\ref{sec:uncertainty:propagation}). We first apply the statistical machinery of inductive conformal prediction (introduced in Section~\ref{sec:pre:icp}), with \emph{nonconformity} functions inspired by the design of residual functions in classical geometric vision~\cite{kahl08tpami-multiple}, to conformalize the heatmaps into circular or elliptical prediction sets --one for each keypoint-- that guarantee coverage of the groundtruth keypoints with a user-specified \emph{marginal} probability (Fig.~\ref{fig:methodoverview}(b)). This provides a simple and general methodology to bound the keypoint prediction errors (\ie, addressing (C1)). 
Given the keypoint prediction sets,
we reformulate the constraints (enforced by the prediction sets) on the keypoints as constraints on the object pose, leading to a \emph{Pose UnceRtainty SEt} (\purse) that guarantees coverage of the groundtruth pose with the same probability. Fig.~\ref{fig:methodoverview}(c) plots the boundary of an example {\purse} (roll, pitch, raw angles for the rotation, and Euclidean coordinates for the translation). The {\purse}, however, is an abstract nonconvex set that does not directly admit estimated poses and uncertainty. Therefore, we develop \emph{RANdom SAmple averaGing} (\ransag) to compute an average pose (Fig.~\ref{fig:methodoverview}(d)) and employ semidefinite relaxations to upper bound the worst-case rotation and translation errors between the average pose and the groundtruth (Fig.~\ref{fig:methodoverview}(e)). This gives rise to the first kind of \emph{computable} worst-case probabilistic error bounds for object pose estimation (\ie, addressing (C3)). Our {\purse} methodology has connections to the framework of \emph{unknown-but-bounded} noise estimation in control theory~\cite{milanese91automatica-optimal}, with special provisions to derive the bounds in a statistically principled way and enable efficient computation.

We test our framework on the LineMOD Occlusion (\lmo) dataset~\cite{brachmann14eccv-linemodocc} to verify the correctness of the theory (Section~\ref{sec:experiments}). First, we empirically show that the {\purse} indeed contains the groundtruth pose according to the user-specified probability. Second, we demonstrate the correctness of the worst-case error bounds: when the {\purse} contains the groundtruth, our bounds are always larger than, and in many cases close to, the actual errors between the average pose and the groundtruth pose. Third, we benchmark the accuracy of the average pose (coming from \ransag) with representative two-stage pipelines based on sparse keypoints (\eg, PVNet~\cite{peng19cvpr-pvnet}) and show that the average pose achieves better or similar accuracy.

{\bf Limitations}. A drawback of our approach, and conformal prediction in general, is that the size of the prediction sets depends on the nonconformity function (whose design can be an art) and may be conservative. 
Our experiments suggest the bounds are loose when the keypoint prediction sets are large (\eg, giving $180^\circ$ rotation bound). We discuss challenges and opportunities in tightening the bounds.


\section{Related Work}
\label{sec:related-work}

{\bf Image-based object pose estimation}. We categorize object pose estimation into two paradigms: \emph{single-stage} and \emph{two-stage}. The latter first detects 2D-3D correspondences and then estimates the object pose via solving a \pnp problem, while the former produces poses without intermediate correspondences. (i) \emph{Single-stage}. Early methods perform pose estimation via template matching~\cite{huttenlocher93pami-comparing,gu10eccv-discriminative,hinterstoisser11pami-gradient}. Recently, deep learning-based approaches such as PoseNet~\cite{kendall15iccv-posenet} and PoseCNN~\cite{xiang18rss-posecnn} applied CNNs to directly regress poses. A major challenge of pose regression is the nonlinearity of 3D rotations, and motivated formulating regression as classification~\cite{su15iccv-render,tulsiani15cvpr-viewpoints,sundermeyer18eccv-implicit} or designing better rotation representations~\cite{zhou19cvpr-continuity,labbe20eccv-cosypose}. It is also popular to predict multiple pose hypotheses followed by voting~\cite{liebelt08cvpr-independent,sun10eccv-depth,michel17cvpr-global}. (ii) \emph{Two-stage}. Early research used handcrafted features~\cite{lowe99iccv-sift,rothganger06ijcv-3d,lepetit05cgv-monocular} to establish 2D-3D correspondences and focused on developing algorithms for solving {\pnp}. Notable algorithms include the minimal solver \pthreep~\cite{gao03pami-p3p,kneip11cvpr-p3p} and variants of the nonminimal solver \pnp~\cite{kneip14eccv-upnp,lepetit09ijcv-epnp,olsson06icpr-optimal,yang20cvpr-perfect}. Outliers (\ie, wrong correspondences) motivated robust estimation based on \ransac~\cite{fischler81acm-ransac}, graduated non-convexity~\cite{yang20ral-gnc,black96ijcv-unification,blake87book-visual}, branch-and-bound~\cite{jiao20iros-globally,li09iccv-consensus,campbell17iccv-globally}, or semidefinite relaxations~\cite{yang22pami-certifiably}. Unreliable correspondences soon became the bottleneck and learned correspondences have been predominant. Learned correspondences can be \emph{sparse} or \emph{dense}. Sparse methods define a handful of keypoints
and predict locations of the keypoints via direct regression~\cite{rad17iccv-bb8,tekin18cvpr-yolo}, probabilistic heatmap~\cite{pavlakos17icra-semantic,oberweger18eccv-heatmap}, or voting~\cite{peng19cvpr-pvnet}. Dense methods~\cite{brachmann16cvpr-uncertainty,li19iccv-cdpn,zakharov19iccv-dpod,park19iccv-pix2pose,hodan20cvpr-epos,wang21cvpr-gdrnet} regress for each object pixel the coordinates of its corresponding 3D point.
Recent literature focus on end-to-end training via differentiating {\pnp}~\cite{brachmann18cvpr-learning,chen20cvpr-backproppnp,campbell20eccv-blindpnp,iwase21iccv-repose,chen22cvpr-epropnp}. Both single-stage and two-stage methods perform well on standard benchmarks~\cite{hodan18eccv-bop}, but a crucial feature that is missing, especially when deploying computer vision algorithms in safety-critical applications, is that these methods do not provide \emph{provably correct} uncertainty quantification and \emph{formal} error bounds \wrt the groundtruth (for either the correspondences or the poses). In this paper, we provide rigorous guarantees by applying conformal prediction to an existing keypoint detection method (the heatmap~\cite{pavlakos17icra-semantic}) and leveraging old and new techniques in computer vision to derive formal error bounds. 

{\bf Conformal prediction in computer vision}. Conformal prediction~\cite{vovk05book-conformal} is a statistical machinery that offers provably correct finite-sample uncertainty quantification without assumptions on the data distribution or the prediction model (\ie, offering a set prediction, instead of a point prediction, that guarantees probabilistic coverage of the groundtruth). \emph{Inductive conformal prediction}~\cite{papadopoulos08chapter-icp} is the most popular variant of conformal prediction because it does not require retraining of the prediction models~\cite{lei13jasa-distribution,angelopoulos21arxiv-gentle,angelopoulos21iclr-conformal}.
Applying conformal prediction to computer vision, however, is still in its infancy. Existing works focus on image classification~\cite{romano20neurips-classification,angelopoulos21iclr-conformal}, tumor segmentation~\cite{wieslander20jbhi-tumor,angelopoulos22arxiv-conformalriskcontrol,bates21jacm-rcps}, and bounding box detection~\cite{li22arxiv-towards,de22csrs-object,angelopoulos21arxiv-gentle}, which are classification or low-dimensional regression problems. Inspired by these works, our unique contributions in this paper are: (i) we apply conformal prediction to keypoints detection, a high-dimensional regression problem; (ii) we design new nonconformity functions and discuss their connections with classical geometric vision; and (iii) we develop algorithms that propagate the uncertainty after conformal prediction to form prediction sets of 6D poses, which are nonlinear and nonconvex manifold objects.

{\bf Performance guarantees}. Pose estimation from 2D-2D, 2D-3D, and 3D-3D correspondences are foundational problems in computer vision textbooks~\cite{hartley03book-geometry,barfoot17book-state,ma04book-invitation,szeliski22book-computer} and typically boil down to formulating and solving mathematical optimization problems.
Benchmarking on simulated and real datasets has been a widely adopted standard for testing different formulations and solvers. However, empirical performance can be misleading without theoretical guarantees. A striking fact is that, though error analysis is an important topic in applied math~\cite{candes06cpam-stable,klivans18colt-efficient,diakonikolas22arxiv-list} and control theory~\cite{milanese91automatica-optimal,soderstrom07automatica-errors,mazzaro04cdc-set}, there is very limited literature in computer vision that reason about \emph{worst-case estimation errors} between the optimal solution and the groundtruth. A popular heuristic relies on the inverse of the Hessian at an optimal solution, which provides the \emph{Cramer-Rao lower bound} on the covariance of the solution (for linear regression this coincides with the covariance)~\cite[Section B.6]{szeliski22book-computer} and thus cannot \emph{upper bound} the estimation errors. Recent works~\cite{rosen19ijrr-sesync,yang20tro-teaser,carlone22arxiv-estimation} derived error bounds for a few geometric vision problems. However, the bounds either depend on uncheckable assumptions and cannot be computed~\cite{rosen19ijrr-sesync,yang20tro-teaser}, or build on machinery (\eg, sum-of-squares proof~\cite{moitra20book-sos,barak16course-proofs}) that only applies to estimators based on moment relaxations~\cite{carlone22arxiv-estimation}, which are still computationally expensive in practice~\cite{yang22mp-inexact}. In this paper, we develop the first kind of efficiently computable error bounds that only require the assumption of \emph{exchangeability} (which comes from conformal prediction). We justify this assumption on our test dataset and numerically show our bounds can be tight for a subset of the test problems.


\section{Inductive Conformal Prediction}
\label{sec:pre:icp}
Given a set $\{ z_i = ( x_i, y_i ) \}_{i=1}^l$ with observation $x_i \in \calX$ and label $y_i \in \calY$ such that each $z_i \in \calZ := \calX \times \calY $ is drawn i.i.d. from an \emph{unknown} distribution on $\calZ$, inductive conformal prediction (ICP) provides 
a \emph{set prediction} $\Feps(x) \subseteq \calY$, parameterized by an error rate $0 < \epsilon <1$, such that given a new sample $z_{l+1} = (x_{l+1},y_{l+1})$ satisfying an \emph{exchangeability} condition (elaborated in Theorem~\ref{thm:icp-validity}), we have
\bea\label{eq:icpmiscoverage}
\probof{ y_{l+1} \in \Feps(x_{l+1}) } \geq 1-\epsilon, 
\eea
\ie, the prediction set $\Feps$ guarantees to contain the true label $y_{l+1}$ with probability at least $1-\epsilon$. 


{\bf Training}. We start by dividing the dataset into a \emph{proper training set} $\{ z_1,\dots,z_m \}$ and a \emph{calibration set} $\{ z_{m+1},\dots,z_{l} \}$. We shorthand $n = l - m$ as the size of the calibration set.
We learn a prediction function $f: \calX \rightarrow \tcalY$ from the proper training set using \emph{any} architecture, which allows us to fully exploit the power of modern deep learning. The prediction space $\tcalY$ can be the same as the label space $\calY$, or can contain auxiliary information such as a heuristic notion of uncertainty (\eg, softmax scores in classification or a heatmap in the case of keypoint detection). 

{\bf Conformal calibration}. 
We define a \emph{nonconformity} function $S: \calZ^{m} \times \calZ \rightarrow \Real{}$ to measure how well a given sample $z = (x,y)$ \emph{conforms} to the proper training set. A popular instance of $S$ leverages the learned prediction $f$:
\bea \label{eq:nonconformity}
S\parentheses{\cbrace{z_1,\dots,z_m},(x,y)} \stackrel{\eg}{=} r(y,f(x)),
\eea
where $r: \calY \times \tcalY \rightarrow \Real{}$ is a measure of disagreement between the label $y$ and the prediction $f(x)$. For example, consider $\calY = \tcalY = \Real{}$, one can design $r(y,f(x)) = \abs{y - f(x)}$: if $(x,y)$ poorly conforms to the training set, $f$ will incur large errors.   
While the function $S$ can be arbitrary (\eg, a learnable neural network~\cite{stutz22iclr-learnconformal}), \eqref{eq:nonconformity} is a convenient definition since $f$ is implicitly dependent on $\{z_i\}_{i=1}^m$ and $r$ can incorporate domain-specific knowledge.
We then compute the nonconformity scores on the calibration set as $\alpha_i = r(y_i,f(x_i)), i = m+1,\dots,l$,
and sort them in \emph{nonincreasing} order $\alpha_{\pi(1)}\geq\dots \geq \alpha_{\pi(n)}$, where $\pi(i) \in \{m+1,\dots,l\}$ is an index permutation.

{\bf Conformal prediction}. Given a new observation $x_{l+1}$ (with an unknown $y_{l+1}$) and a user-specified $\epsilon \in (0,1)$, we compute the inductive conformal prediction (ICP) set as
\bea\label{eq:icpcompute}
\Feps \parentheses{x_{l+1}} = \cbrace{y \in \calY \mid \alpha^y \leq \alpha_{\pi(\floor{(n+1)\epsilon})}},
\eea
where $\alpha^y = r(y,f(x_{l+1}))$
is the nonconformity score of the new sample when fixing the true label to be $y$. In other words, the ICP set~\eqref{eq:icpcompute} outputs the set of all labels that make the nonconformity score of the new sample no greater than $\alpha_{\pi(\floor{(n+1)\epsilon})}$ -- the $\floor{(n+1)\epsilon}$-th largest nonconformity score in the calibration set. 
We have the following result stating the probabilistic coverage of the ICP set~\eqref{eq:icpcompute}.

\begin{figure}
\vspace{-4mm}
\begin{center}
\includegraphics[width=0.6\columnwidth]{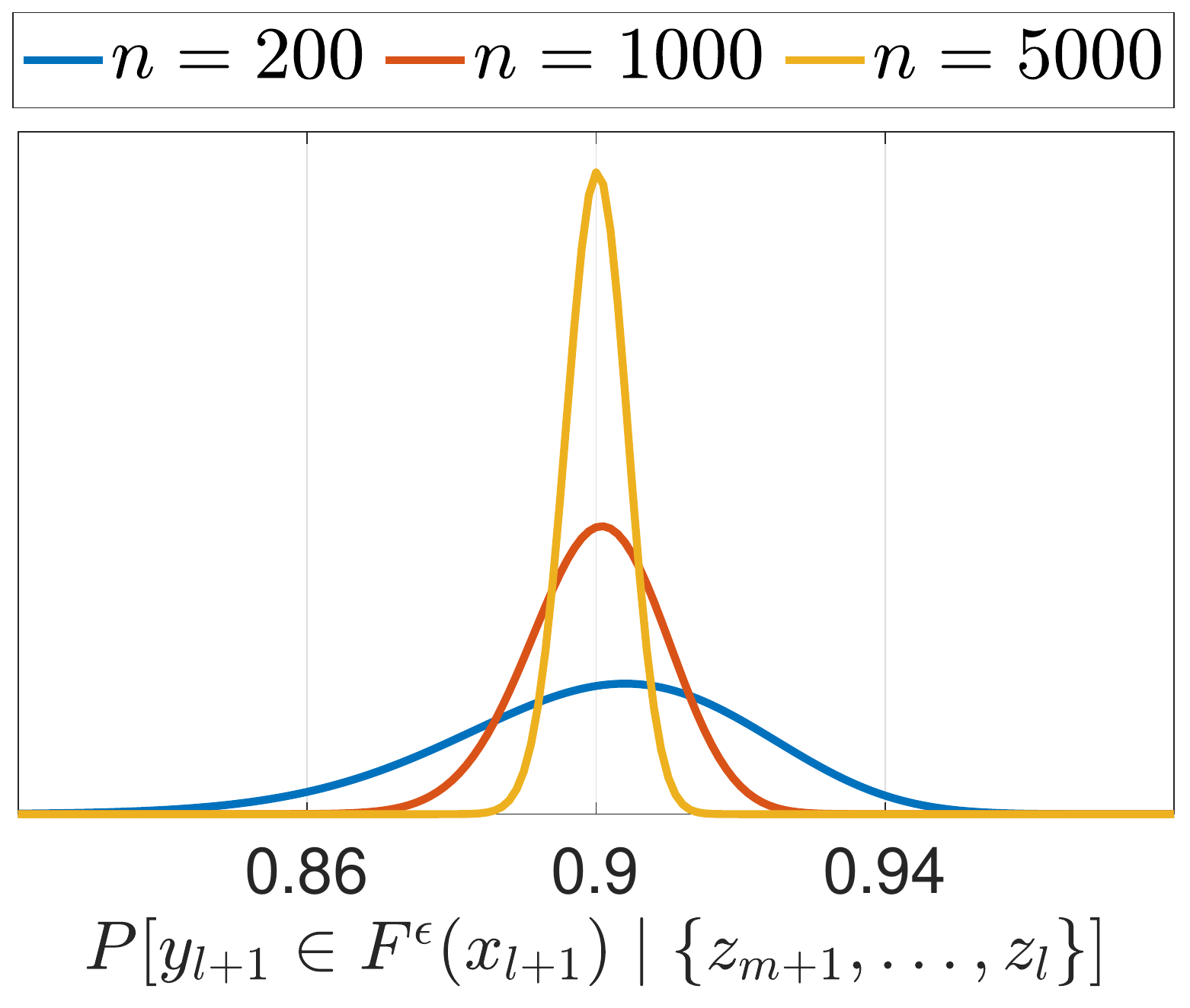}
\end{center}
\vspace{-6mm}
\caption{Beta distribution of the conditional coverage in~\eqref{eq:beta} with $\epsilon=0.1$ and different $n$. Notice how the conditional probability becomes more concentrated around $1-\epsilon$ when $n$ increases.
\label{fig:beta-distribution}}
\vspace{-7mm}
\end{figure}

\begin{theorem}[Validity of ICP Coverage {\cite{vovk05book-conformal,lei18jasa-conformal,vovk12acml-icpconditional}}] \label{thm:icp-validity}
If $z_{m+1},\dots,z_l$, $z_{l+1} = (x_{l+1},y_{l+1})$ are exchangeable, \ie, their distribution is invariant under permutation, then
\bea\label{eq:icpvalidity}
1 - \epsilon \leq \probof{y_{l+1} \in \Feps(x_{l+1})} \leq 1 - \epsilon + 1/(n+1)
\eea
for any $\epsilon \in (0,1)$. Furthermore, when conditioned on the calibration set, calling $h = \floor{(n+1)\epsilon}$, we have
\begin{equation}\label{eq:beta}
\hspace{-4mm}\probof{y_{l+1}\!\in\!\Feps(x_{l+1})\!\mid\!\{z_{m+1},\dots,z_l\}}\!\sim\!\mathrm{Beta}(n+1\!-\!h,h).
\end{equation}
\end{theorem}
A few remarks are in order about Theorem~\ref{thm:icp-validity}.
First, asking $z_{m+1},\dots,z_l,z_{l+1}$ to be exchangeable is weaker than asking them to be independent. However, this assumption typically fails when the calibration set is a single video sequence, where the image frames $\{z_{m+1},\dots,z_l\}$ are temporally correlated~\cite{luo21arxiv-conformalsafety}. Fortunately, as we detail in Section~\ref{sec:experiments}, the way the LineMOD Occlusion dataset~\cite{brachmann14eccv-linemodocc} was collected makes the exchangeability condition easily satisfied, which also suggests best practices to make the exchangeability condition hold in computer vision. 
Second, the lower bound in~\eqref{eq:icpvalidity} can be intuitively proved because under exchangeability, $\alpha_{l+1} := r(y_{l+1},f(x_{l+1}))$ --the nonconformity score of the new sample with the true label-- is \emph{exchangeable} with the nonconformity scores of the calibration samples, and hence \emph{equally likely} to fall in anywhere between the scores $\{ \alpha_{\pi(i)}\}_{i=1}^n$. Consequently, $\probof{y_{l+1} \in \Feps(x_{l+1})} = \probof{\alpha_{l+1} \leq \alpha_{\pi(\floor{(n+1)\epsilon})}} = 1 - \floor{(n+1)\epsilon}/(n+1) \geq 1 - \epsilon$. The upper bound in \eqref{eq:icpvalidity} states that $1-\epsilon$ is not overly conservative (indeed tight if $n$ is large). 
Lastly, the probabilistic guarantee in \eqref{eq:icpvalidity} is \emph{marginal} over the randomness of the calibration set, meaning if one chooses an infinite number of calibration sets,  the \emph{average} empirical coverage will converge to $1-\epsilon$. This, however, implies that the empirical coverage given one calibration set is a random variable that fluctuates as the Beta distribution~\eqref{eq:beta}. Fig.~\ref{fig:beta-distribution} plots the Beta distribution at $\epsilon=0.1$ with different sizes of the calibration set. We observe that as $n$ increases the empirical coverage becomes more concentrated at $1-\epsilon$. Our experiments show that even with a small ($n=200$) calibration set, the empirical coverage is close to, and mostly higher than, $1-\epsilon$.

\section{Conformal Keypoint Detection}
\label{sec:keypoint:conformal}
In this section, we apply the ICP framework in Section~\ref{sec:pre:icp} to the problem of semantic keypoint detection. 

{\bf Setup}. Denote by $x \in \Real{H\times W\times 3}$ an RGB image picturing an object, by $\vy = (y_1,\dots,y_K) \in \Real{2} \times \dots \times \Real{2} := \calY$ the groundtruth locations of $K$ semantic keypoints of the object. We partition a given dataset $\{z_{i}:=(x_{i},\vy_{i})\}_{i=1}^l$ 
into a proper training set (of size $m$) and a calibration set (of size $n$).
We follow the three steps in Section~\ref{sec:pre:icp} to perform ICP.

{\bf Training}. We choose 
the heatmap approach in~\cite{pavlakos17icra-semantic,schmeckpeper22jfr-semantic} as the prediction function: given an image $x$, \cite{schmeckpeper22jfr-semantic} outputs a set of heatmaps $\vf(x) = (f(x)_1,\dots,f(x)_K)$, where each $f(x)_{k} \in \Delta^{HW}:=\{v \in \mathbb{R}^{HW}_{+}\mid \sum_{i}^{HW} v_i = 1\}$ predicts the probability distribution of the $k$-th keypoint lying on each pixel of the image.\footnote{The heatmap in the original paper~\cite{pavlakos17icra-semantic} is not a valid probability distribution as it contains negative values and do not sum up to $1$. We remove the negative values and normalize it to be a valid probability distribution.} For convenience, we use $q^j \in \Real{2}$ to denote the $j$-th pixel location in $x$ and $f(x)_k^j \in \mathbb{R}_{+}$ to denote the probability of the $k$-th keypoint lying on $q^j$. Let $\sigma_k$ be the index permutation that sorts $f(x)_k$ in nonincreasing order, \ie, $f(x)_k^{\sigma_k(1)} \geq \dots \geq f(x)_k^{\sigma_k(HW)}$. As we will soon show, choosing the heatmap approach leads to simple and intuitive designs of the nonconformity function.

{\bf Conformal calibration}. 
We design the following nonconformity function 
\bea\label{eq:maxnonconformity}
r(\vy,\vf(x)) = \max\{ \phi(y_k,f(x)_k) \}_{k=1}^K
\eea
that uses $\phi$ to score each keypoint and then selects the maximum score. This design considers the worst keypoint detection performance of $\vf$. We provide two designs of $\phi$ below.

{\emph{(a) Peak}}. Shorthand $p_k = f(x)_k^{\sigma_k(1)}$ as the peak probability in the $k$-th heatmap and $q_k = q^{\sigma_k(1)}$ as the pixel location attaining the peak probability, we design
\begin{equation}\label{eq:peak}
\phipeak(y_k, f(x)_k) = p_k \norm{y_k - q_k} \tag{peak}
\end{equation}
which computes the error between the true keypoint location $y_k$ and the most probable keypoint location $q_k$ and scales the error by the peak probability $p_k$. $\phipeak$ describes nonconformity because it becomes larger when the network $\vf$ is \emph{confidently wrong} (both $\norm{y_k - q_k}$ and $p_k$ are large), implying the sample is highly nonconforming. 

{\emph{(b) Covariance}}. Let $\barq_k = \sum_{j=1}^J f(x)_k^{\sigma_k(j)} q^{\sigma_k(j)}$ be the expected location of the top-$J$ most likely detections for the $k$-th keypoint, and $\Sigma_k = \sum_{j=1}^J f(x)_k^{\sigma_k(j)} \cdot (q^{\sigma_k(j)} - \barq_k)(q^{\sigma_k(j)} - \barq_k)\tran$ as the covariance, we design
\begin{equation}\label{eq:cov}
\phicov(y_k, f(x)_k) = (y_k - \barq_k)\tran \Sigma_k\inv (y_k - \barq_k) \tag{cov}
\end{equation}
which computes the squared Mahalanobis distance~\cite{mahalanobis36nisi-mahalanobis} from the groundtruth $y_k$ to the top-$J$ keypoint detections (represented by the mean $\barq_k$ and covariance $\Sigma_k$).\footnote{We only choose the top-$J$ ($J=100$) most likely detections on the heatmap because the heatmap can be quite noisy in practice.} A larger Mahalanobis distance indicates more abnormality of the heatmap $f(x)_k$ (compared to the groundtruth $y_k$)~\cite{geun00cs-multivariate}, and hence implies higher nonconformity.

Using the nonconformity function \eqref{eq:maxnonconformity} with \eqref{eq:peak} or \eqref{eq:cov}, we compute the nonconformity scores of the calibration set and sort them as: $\alpha_{\pi(1)} \geq \dots \geq \alpha_{\pi(n)}$.

{\bf Conformal prediction}. Given an error rate $\epsilon \in (0,1)$, we first find $\alpha_{\pi(\floor{(n+1)\epsilon})}$. Then, according to the ICP set definition~\eqref{eq:icpcompute} and our nonconformity function~\eqref{eq:maxnonconformity}, we output the ICP set for a new $x_{l+1}$ as
\bea\label{eq:generalkpticp}
 & \Feps(x_{l+1}) \nonumber \\
\hspace{-4mm} = & \hspace{-3mm} \{\vy \in \calY \mid \max\{ \phi(y_k,f(x_{l+1})) \}_{k=1}^K \leq \alpha_{\pi(\floor{(n+1)\epsilon})} \}  \nonumber \\
\hspace{-4mm} = & \{\vy \in \calY \mid \phi(y_k,f(x_{l+1})) \leq \alpha_{\pi(\floor{(n+1)\epsilon})}, \forall k \},
\eea
where we used $\max\{\phi_1,\dots,\phi_K\} \leq \alpha$ if and only if $\phi_k \leq \alpha$ for any $k$. Insert~\eqref{eq:peak} into~\eqref{eq:generalkpticp}, we have $\Fepsball(x_{l+1})$ as
\begin{equation}\label{eq:icp-ball} \tag{ball}
\cbrace{ \vy \in \calY \mid \norm{y_k - q_{l+1,k}} \leq \frac{ \alpha_{\pi(\floor{(n+1)\epsilon})} }{p_{l+1,k}}, \forall k},
\end{equation}
which defines --for the $k$-th keypoint-- a ball centered at $q_{l+1,k}$ (the most likely detection) with a radius inversely proportional to $p_{l+1,k}$ and proportional to $\alpha_{\pi(\floor{(n+1)\epsilon})}$. Similarly, insert~\eqref{eq:cov} into~\eqref{eq:generalkpticp}, we have $\Fepsellipse(x_{l+1})$ as
\begin{equation}\label{eq:icp-ellipse} \tag{ellipse}
\hspace{-4mm}\cbrace{ \vy \in \calY \mid (y_k - \barq_{l+1,k})\tran \frac{\Sigma_{l+1,k}\inv }{\alpha_{\pi(\floor{(n+1)\epsilon})}} (y_k - \barq_{l+1,k}) \leq 1, \forall k},
\end{equation}
which defines --for the $k$-th keypoint-- an ellipse centered at $\barq_{l+1,k}$ (the expected location of the top-$J$ detections) with an area proportional to $\det(\Sigma_{l+1,k})$ and $\alpha_{\pi(\floor{(n+1)\epsilon})}$.\footnote{The area of  $(x-\mu)\tran A (x-\mu) \leq 1$ is proportional to $\det(A\inv)$.} From \eqref{eq:icp-ball} and \eqref{eq:icp-ellipse}, we observe that the prediction sets become larger when (i) the heatmaps are uncertain, \ie, the peak probability is low or the covariance matrix has large determinant; and (ii) the heatmaps perform poorly on the calibration set, leading to a large $\alpha_{\pi(\floor{(n+1)\epsilon})}$.

{\bf Connections to geometric vision}. Our nonconformity function bears similarity to the \emph{residual} function in geometric vision~\cite{hartley03book-geometry,antonante21tro-outlier,chin18eccv-robust}. For example, the \eqref{eq:peak} and \eqref{eq:cov} functions are similar to the (weighted) reprojection error~\cite{hartley03book-geometry}, and the ``$\max$'' in~\eqref{eq:maxnonconformity} can be connected to seminal work on optimizing the $\ell_{\infty}$ norm~\cite{kahl08tpami-multiple}.

{\bf Outlier-robust nonconformity}? One potential issue of the nonconformity function~\eqref{eq:maxnonconformity} is that a \emph{single} {outlier} can inflate the score and the calibration quantile $\alpha_{\pi(\floor{(n+1)\epsilon})}$ and lead to conservative prediction sets (\eg, when $\vf$ predicts $K-1$ keypoints perfectly but misses one keypoint). A potential remedy in geometric vision is to use robust cost functions~\cite{black96ijcv-unification,yang20ral-gnc,barron19cvpr-general}. Therefore, a natural question is whether ``robustifying'' the nonconformity function \eqref{eq:maxnonconformity} can lead to better prediction sets. Here we focus on only robustifying $\phi$ in \eqref{eq:maxnonconformity} and provide a negative answer.

\begin{proposition}[Invariance of ICP]\label{prop:invariance}
Let $\rho: \mathbb{R}_+ \mapsto \mathbb{R}_+$ be any monotonically increasing function. Fixing the calibration set and error rate $\epsilon$, the nonconformity function
\bea\label{eq:robustnonconformity}
r_{\rho}(\vy, \vf(x)) = \max\{ \rho(\phi(y_k,f(x)_k)) \}_{k=1}^K
\eea
leads to the same ICP set as \eqref{eq:maxnonconformity}.
\end{proposition}
The proof of Proposition~\ref{prop:invariance} is presented in Supplementary Material. We conclude that common robust costs, such as $\ell_1$, Huber, Geman-McClure, and Barron's adaptive kernel~\cite{black96ijcv-unification,barron19cvpr-general} (which are monotonically increasing on $[0,+\infty]$) cannot change the ICP sets by robustifying the individual score $\phi$. However, it remains an open question whether changing the ``$\max$'' operation in \eqref{eq:maxnonconformity} can give rise to better ICP sets. For instance, replacing ``$\max$'' with ``$\sum$'' in \eqref{eq:maxnonconformity} and using the Geman-McClure robust cost $\rho(\phi) = \frac{\phi^2}{1 + \phi^2}$ with $\phi = \phipeak$ results in the following ICP set 
\bea
\cbrace{ \vy \in \calY \mid \sum_{k=1}^K \frac{p_k^2 \norm{y_k - q_k}^2}{1 + p_k^2 \norm{y_k - q_k}^2} \leq \alpha_{\pi(\floor{(n+1)\epsilon})} }
\eea
that does not admit a geometric interpretation that is as simple and intuitive as the \eqref{eq:icp-ball} and \eqref{eq:icp-ellipse} sets introduced before. In fact, it is indeed the simplicity of \eqref{eq:icp-ball} and \eqref{eq:icp-ellipse} that enables us to propagate the uncertainty in keypoints to the object pose, as we will show in the next section.

\section{Geometric Uncertainty Propagation}
\label{sec:uncertainty:propagation}
Conformalizing the heatmaps gives us prediction sets that guarantee probabilistic coverage of the true keypoints. We unify the prediction sets \eqref{eq:icp-ball} and \eqref{eq:icp-ellipse} as
\bea\label{eq:icpunify}
\Feps(x) = \cbrace{\vy \in \calY \mid (y_k - \mu_k)\tran \Lambda_k (y_k - \mu_k) \leq 1,\forall k},
\eea
where $\mu_k = q_{l+1,k}, \Lambda_k = \frac{p_{l+1,k}^2}{\alpha^2_{\pi(\floor{(n+1)\epsilon})}} \eye_2$ for~\eqref{eq:icp-ball}, $\mu_k = \barq_{l+1,k},\Lambda_k = \frac{\Sigma_{l+1,k}\inv}{\alpha_{\pi(\floor{(n+1)\epsilon})}}$ for~\eqref{eq:icp-ellipse}, and we omit the subscript $l+1$ for simplicity.

{\bf Why not uncertainty-aware \pnp}? A popular way to estimate pose from~\eqref{eq:icpunify} is to solve an uncertainty-aware \pnp
\bea \label{eq:uncertainpnp}
\min_{(R,t) \in \SEthree} & \displaystyle \sum_{k=1}^K (y_k - \mu_k)\tran \Lambda_k (y_k - \mu_k) \nonumber \\
\subject & y_k = \Pi(RY_k + t), k=1,\dots,K
\eea
where $Y_k \in \Real{3},k=1,\dots,K$ are the 3D object keypoints and $\Pi(\cdot)$ denotes the camera projection. We challenge this approach and point out its two drawbacks. First, it is difficult to solve~\eqref{eq:uncertainpnp} to global optimality due to (i) the nonconvex $\SEthree$ constraint and (ii) the rational polynomial appearing in $\Pi(\cdot)$. The best known approach to solve \eqref{eq:uncertainpnp} relies on either branch-and-bound~\cite{olsson06icpr-optimal} or local optimization. Second, solving~\eqref{eq:uncertainpnp} typically outputs a \emph{single} optimal pose without uncertainty quantification. Are there other poses that attain similar costs as the optimal pose? How close is the optimal pose to the groundtruth pose? These questions remain not answered in the literature.

{\bf Pose UnceRtainty SEt (\purse)}. We propose to, instead of solving a {\pnp} problem similar to~\eqref{eq:uncertainpnp}, directly propagate the uncertainty in the ICP sets to the object pose. 
\begin{proposition}[\purse]\label{prop:purse}
Let $\sgt = [\vectorize{\Rgt}\tran; \tgt\tran]\tran$ be the groundtruth object pose (that lies in front of the camera). Then, the groundtruth keypoints $\vy = (y_1,\dots,y_K)$ belong to the ICP set $\Feps(x)$ in~\eqref{eq:icpunify} if and only if $\sgt$ belongs to the following pose uncertainty set
\begin{equation}\label{eq:purse}
\hspace{-3mm} \Seps = \cbrace{s \in \SEthree \ \middle\vert\ \substack{ \displaystyle s\tran A_k s \leq 0,k=1,\dots,K \\ \displaystyle b_k\tran s > 0, k=1,\dots,K} }, \tag{\purse}
\end{equation} 
where $A_k\in \sym{12}, b_k \in \Real{12},k=1,\dots,K$ are constant matrices dependent on $\mu_k,\Lambda_k,Y_k$ and camera intrinsics.
\end{proposition}
The detailed proof for Proposition~\ref{prop:purse} is algebraically involved and postponed to the Supplementary Material. The high-level intuition is, however, straightforward: we plug in $y_k = \Pi(RY_k + t)$ into \eqref{eq:icpunify} and obtain $K$ quadratic inequalities of the form $s\tran A_k s \leq 0$. The linear inequalities $b_k\tran s > 0$ are added to enforce the (transformed) 3D keypoints lie in front of the camera. Proposition~\ref{prop:purse} implies, if we are $1-\epsilon$ confident the groundtruth keypoints can be anywhere inside $\Feps(x)$, then we should also be confident any pose in~\eqref{eq:purse} can be the groundtruth. Viewing pose estimation as a set estimation with guaranteed probabilistic coverage of the groundtruth is fundamentally different from viewing it as computing a single pose from~\eqref{eq:uncertainpnp} that is (hopefully) close enough to the groundtruth.

{\bf RANdom SAmple averaGing (\ransag)}. Verifying if a given pose belongs to the \purse is straightforward via checking the inequalities in~\eqref{eq:purse}. However, the {\purse} does not directly give us estimated poses. Therefore, we propose an efficient sampling algorithm called \emph{RANdom SAmple averaGing} (\ransag) that is analogous to {\ransac}~\cite{fischler81acm-ransac} and leverages the minimal solver \pthreep~\cite{gao03pami-p3p}, presented in Algorithm~\ref{alg:ransag}.
The intuition is that, though it is difficult to sample directly in \purse due to the (nonconvex) constraints, it is easy to sample from the keypoint prediction set~\eqref{eq:icpunify} due to its simple geometry (balls and ellipses). Thus, at each iteration (line~\ref{line:p3piter}) {\ransag} samples three keypoints (line~\ref{line:samplek}-\ref{line:samplekeypoints}), solves the \pthreep inverse problem, and accept the poses that belong to the \eqref{eq:purse} (line~\ref{line:solvep3p}). \ransag typically returns around $100$ valid samples with $T=1000$ trials. However, in difficult cases (\eg, when $\Seps$ is small or even empty) it is possible to obtain zero samples ($S = \emptyset$). In this situation, {\ransag} samples $\floor{T/20}$ (default $50$) poses without checking if they belong to the \purse, via sampling $K$ keypoints and solving \pnp (line~\ref{line:emptyset}-\ref{line:solvepnp}).\footnote{Here we switch from \pthreep to \pnp because \pnp uses all $K$ keypoints and there is less ambiguity in its solution.} After obtaining a set of poses, {\ransag} performs rotation averaging (line~\ref{line:rotavg}) and translation averaging (line~\ref{line:transavg}) to obtain an average pose $\bar{s}$.\footnote{In Algorithm~\ref{alg:ransag} we use rotation averaging with the Chordal distance metric. The user is free to choose other single rotation averaging algorithms with different distance metrics~\cite{hartley13ijcv-rotation}.} Note that {\ransag} does not check if $\bar{s}$ lies in the \purse.

\setlength{\textfloatsep}{5pt}%
\begin{algorithm}[t]
\SetAlgoLined
{\bf Input:} an ICP set $\Feps(x)$~\eqref{eq:icpunify} and its corresponding \eqref{eq:purse} $\Seps$; maximum trials $T$; initial $\hat{S} = \emptyset$; \\
{\bf Output:} sample poses ${S} \subset \SEthree$ in \purse, and an average pose $\bar{s} \in \SEthree$; \\
\For{ $\tau \gets 1$ to $T$ \label{line:p3piter}}{
	Sample $\{k_1,k_2,k_3\}$ from $[K]$ ($k_1 \neq k_2 \neq k_3$); \label{line:samplek}\\
	Sample $\haty_{k_i}, i=1,2,3$ from \label{line:samplekeypoints}
	\begin{equation}
	\{ y \in \Real{2} \mid (y - \mu_{k_i})\tran \Lambda_{k_i} (y - \mu_{k_i}) \leq 1 \}; \nonumber
	\end{equation}\\
	$\hat{S} \leftarrow \hat{S} \cup ( \Seps \cap \text{\pthreep}(\{\haty_{k_i} \leftrightarrow Y_{k_i}\}_{i=1}^3 ) )$; \label{line:solvep3p}
}
$S = \hat{S}$;\\
\If{$\hat{S} = \emptyset$ \label{line:emptyset} }{
	\For{$\tau \gets 1$ to $\floor{T/20}$}{
	Sample $\haty_{k}, k=1,\dots,K$ from $\Feps(x)$;\\
	$\hat{S} \leftarrow \hat{S} \cup \text{\pnp}(\{\haty_{k} \leftrightarrow Y_{k}\}_{k=1}^K)$;\label{line:solvepnp}
	}
}
\label{line:rotavg} $\bar{R} = \text{proj}_{\SOthree}( \sum_{(R_j,*) \in \hat{S}} R_j )$; \\
\label{line:transavg} $\bar{t} = \frac{1}{\vert \hat{S} \vert } \sum_{(*,t_j) \in \hat{S}} t_j $; \\
{\bf return:} $S$, $\bar{s} = (\bar{R},\bar{t})$ 
\caption{RANdom SAmple averaGing \label{alg:ransag}}
\end{algorithm}

{\bf Worst-case error bounds}. To upper bound the errors between the average pose $\bar{s}$ and the groundtruth $(\Rgt,\tgt)$, we maximize the squared \emph{pose-to-\purse} distance:
\begin{equation}\label{eq:pose2purse}
d_{\epsilon,\lambda}^2 = \max_{(R,t) \in \Seps} \lambda \Fnorm{R - \bar{R}}^2 + (1-\lambda) \norm{t - \bar{t}}^2
\end{equation} 
given $\lambda \in [0,1]$. Particularly, we compute two cases $\lambda = 1$ (the maximum rotation distance) and $\lambda = 0$ (the maximum translation distance). Proposition~\ref{prop:purse} states the groundtruth $(\Rgt,\tgt)$ lies in $\Seps$ with $1-\epsilon$ probability, hence
\bea\label{eq:boundRt}
\Fnorm{\bar{R} - \Rgt} \leq d_{\epsilon,1}, \quad \norm{\bar{t} - \tgt} \leq d_{\epsilon,0}
\eea
holds with probability $1-\epsilon$.

{\bf Computing the bounds}. Problem~\eqref{eq:pose2purse} is nonconvex due to the constraints of the \eqref{eq:purse} $\Seps$. 
We relax the nonconvex problem~\eqref{eq:pose2purse} into a convex semidefinite program (SDP) and employ off-the-shelf solvers to optimize the SDP~\cite{yang22pami-certifiably,briales18cvpr-certifiably,kahl07ijcv-globally}.\footnote{We omit the technical details and refer the interested reader to \cite[Section 2]{yang22pami-certifiably} for a pragmatic introduction to SDP relaxations. In practice, we use the code provided by~\cite{yang22pami-certifiably} in \url{https://github.com/MIT-SPARK/CertifiablyRobustPerception}, apply a second-order SDP relaxation to~\eqref{eq:pose2purse}, and use MOSEK~\cite{mosek} to solve the SDP (in about 8 seconds). Solving a first-order SDP relaxation of~\eqref{eq:pose2purse} takes about $0.1$ second but yields looser bounds.} Two possible outcomes can happen: (i) the optimal SDP value coincides with the optimal value of~\eqref{eq:pose2purse}. The relaxation is said to be \emph{exact} and one can extract an optimal solution of~\eqref{eq:pose2purse} from the SDP, or (ii) the relaxation is not exact, but the optimal SDP value still provides an \emph{upper bound} for the optimal value of~\eqref{eq:pose2purse}. Therefore, we either exactly compute $d^2_{\epsilon,\lambda}$ or find an upper bound, both can bound the worst-case error (\cf~\eqref{eq:boundRt}).\footnote{The \purse can potentially be empty, leading to infeasibility of problem~\eqref{eq:pose2purse}. In such cases, empirically the SDP solver returns ``\texttt{PRIMAL\_INFEASIBLE}'' (red squares lying on the $y$-axis of Fig.~\ref{fig:coverage-and-bound}).}


We end with a remark about computing tighter bounds.

\begin{remark}[Best Worst-case Error Bounds]
\label{rmk:bestbound}
\eqref{eq:pose2purse} can be used to bound errors for all possible pose estimators (\eg, from \pnp~\eqref{eq:uncertainpnp}). What is the best estimator that attains the smallest error bounds? This boils down to solving
\bea\label{eq:bestbound}
\min_{(\bar{R},\bar{t}) \in \SEthree} \left[ \max_{(R,t)\in \Seps} \lambda \Fnorm{R - \bar{R}}^2 + (1-\lambda)\norm{t - \bar{t}}^2\right]
\eea
whose solution is known as the \emph{Chebyshev center}~\cite{milanese91automatica-optimal,eldar08sp-minimax} of the \purse $\Seps$. Unfortunately, problem~\eqref{eq:bestbound} is more challenging than~\eqref{eq:pose2purse} and there is no efficient algorithm to solve it to global optimality. In the Supplementary Material, we evaluate the worst-case error bounds for multiple $(\bar{R},\bar{t})$ samples, select the smallest bounds, and compare them with those of the average pose. An interesting future research direction is to explore differentiable optimization~\cite{pineda22neurips-theseus} or bilevel polynomial optimization~\cite{nie17siopt-bilevel} to solve~\eqref{eq:bestbound}.
\end{remark}
\vspace{-4mm}


\newcommand{\coverwidth}{5.4cm}
\newcommand{\boundwidth}{4.2cm}
\begin{figure*}
\vspace{-8mm}
\begin{center}
\begin{minipage}{\textwidth}
\centering
\begin{tabular}{ccc}%
	    \begin{minipage}{\coverwidth}%
		\centering%
		\includegraphics[width=\columnwidth]{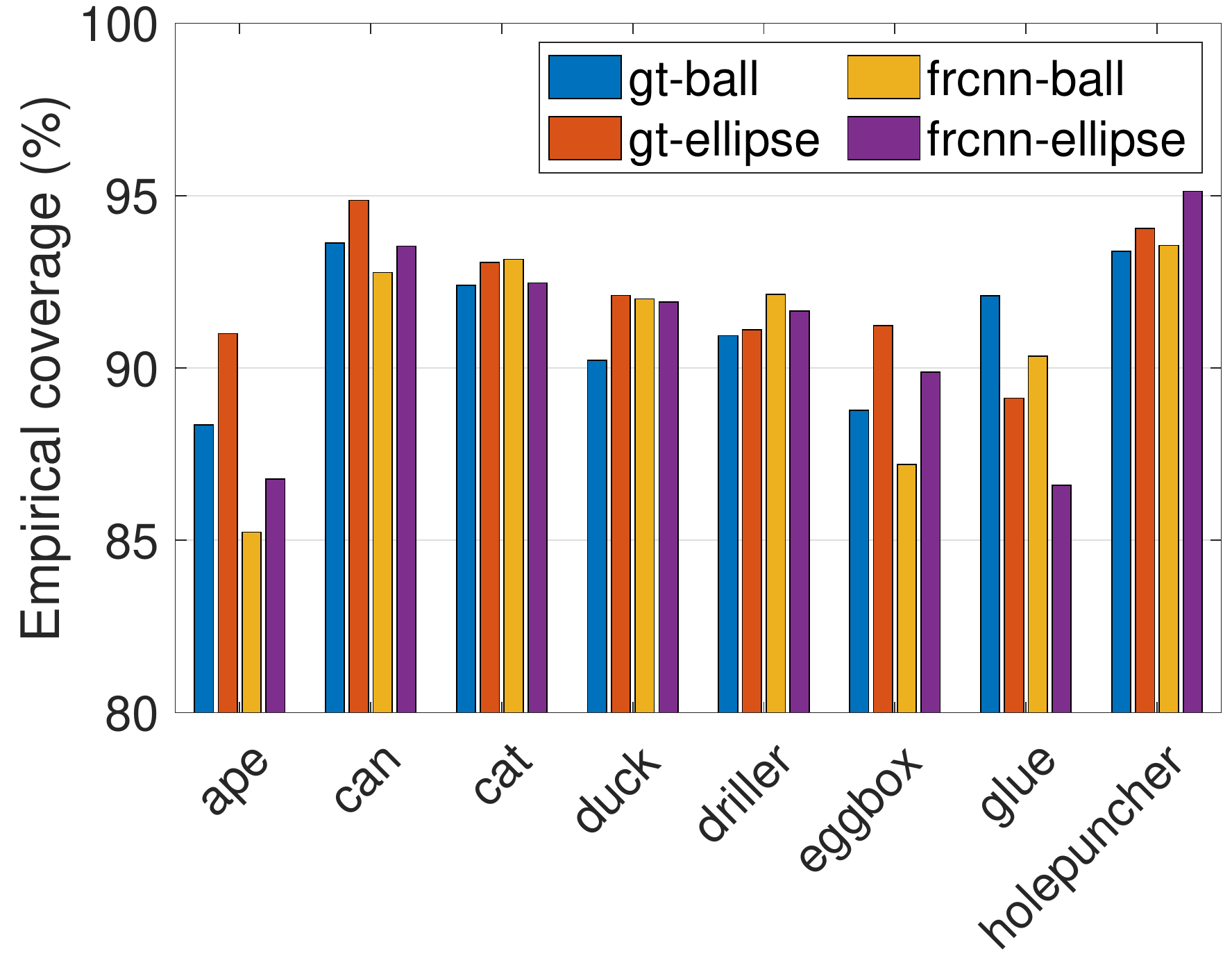}
		\end{minipage}
	&  
	    \begin{minipage}{\boundwidth}%
		\centering%
		\includegraphics[width=\columnwidth]{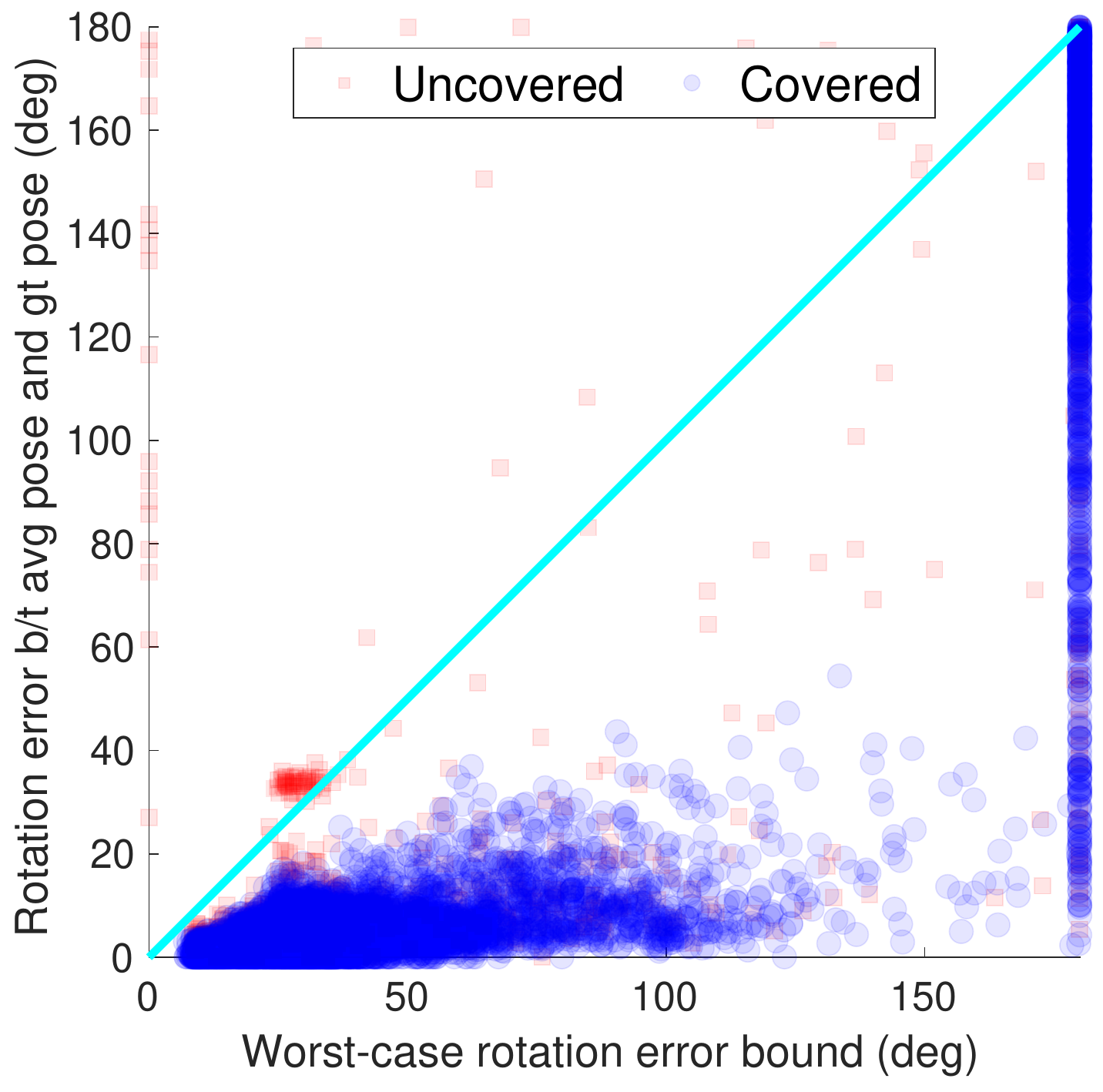}
		\end{minipage}
	&  	
	    \begin{minipage}{\boundwidth}%
		\centering%
		\includegraphics[width=\columnwidth]{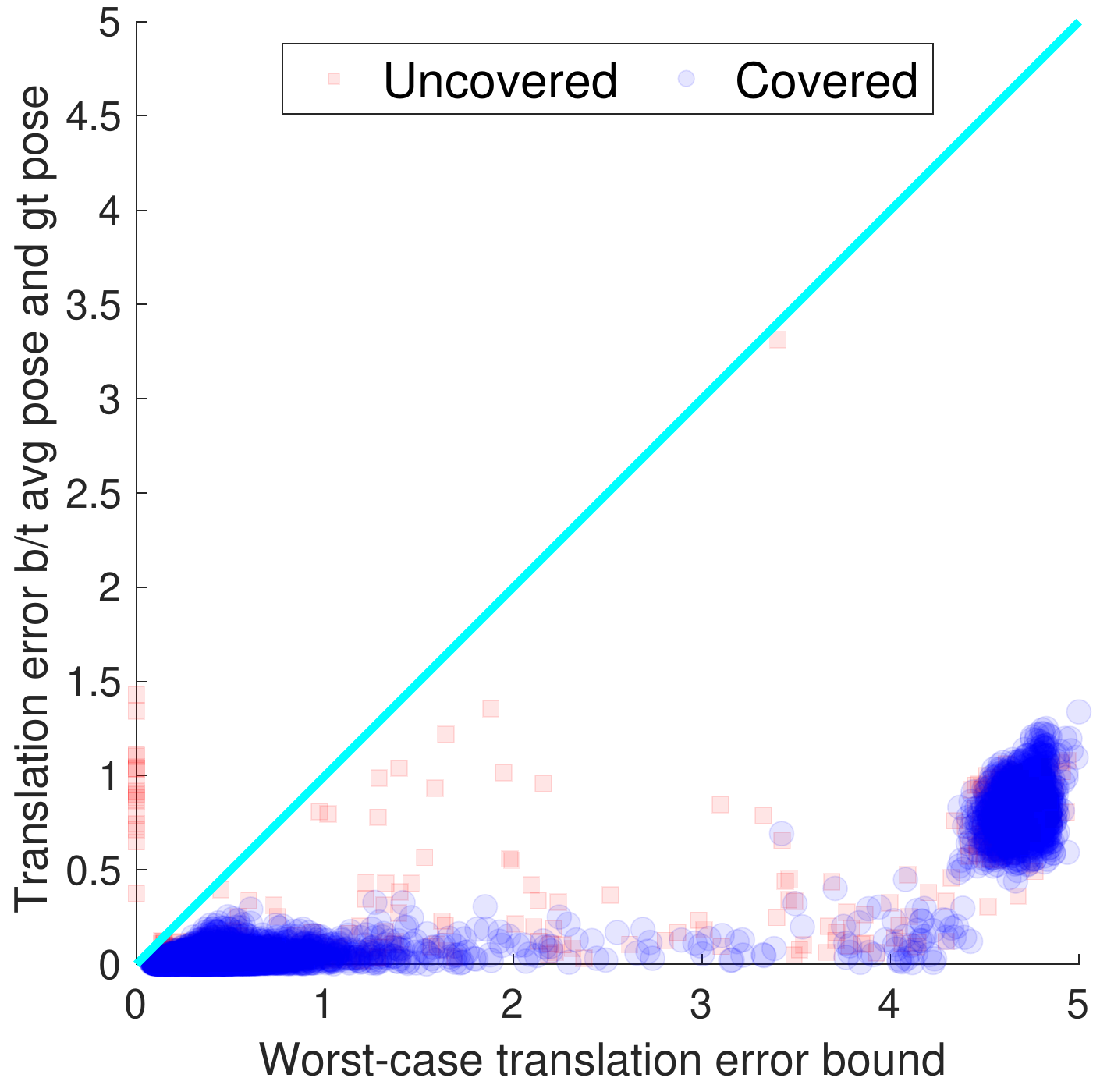}
		\end{minipage}
	\\
		\begin{minipage}{\coverwidth}%
		\centering%
		\includegraphics[width=\columnwidth]{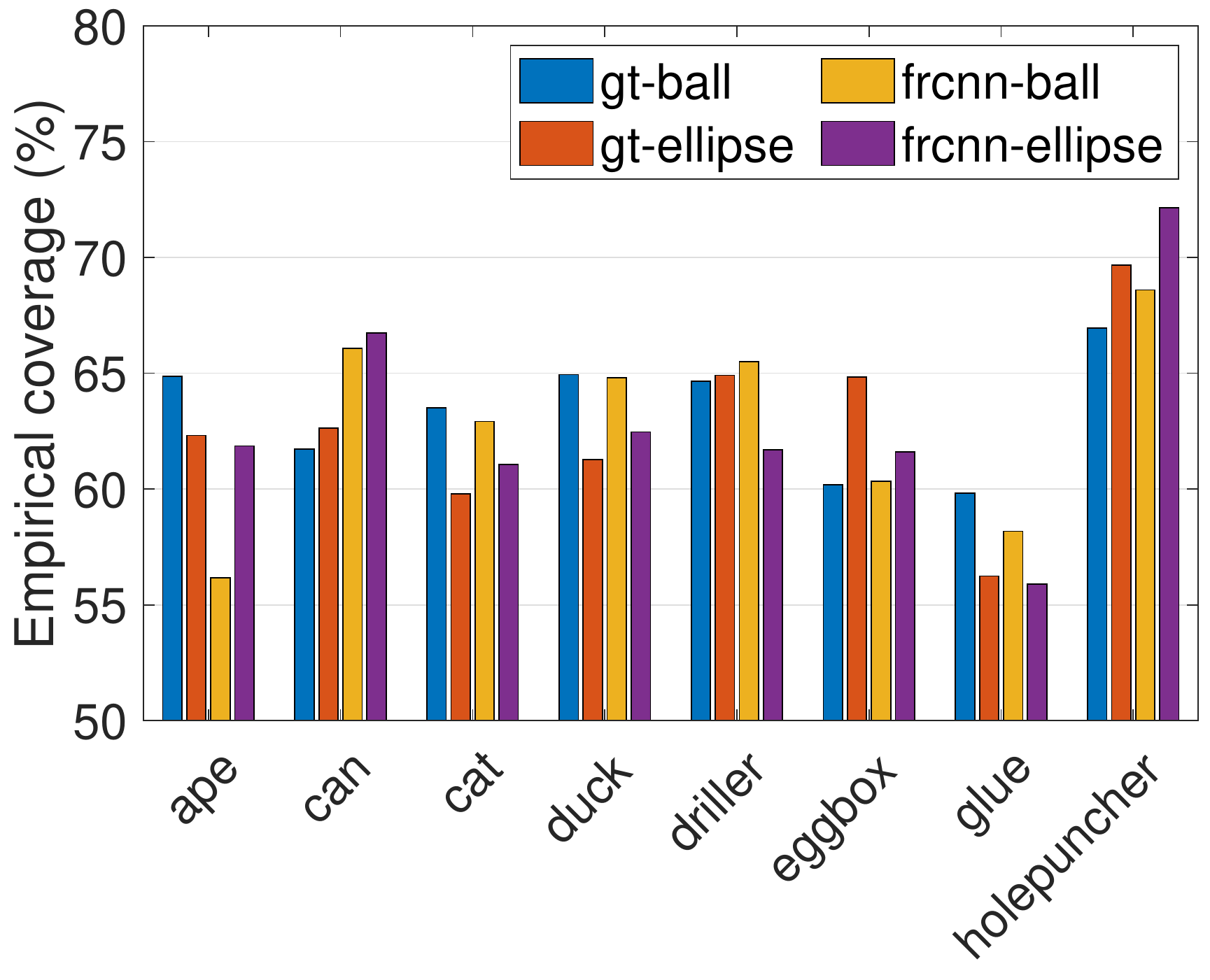}
		\end{minipage}
	& 
	    \begin{minipage}{\boundwidth}%
		\centering%
		\includegraphics[width=\columnwidth]{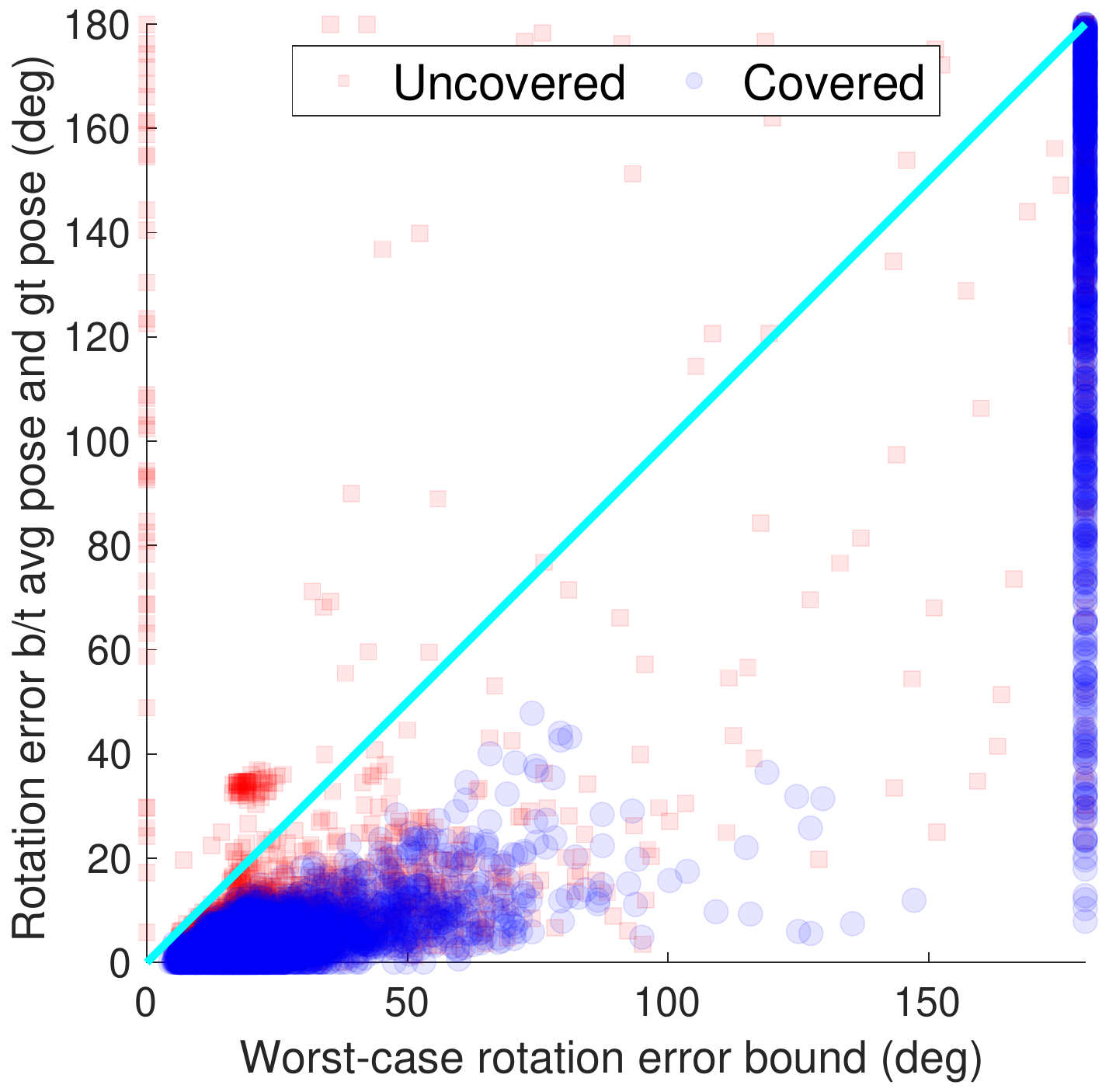}
		\end{minipage}
	&  
	    \begin{minipage}{\boundwidth}%
		\centering%
		\includegraphics[width=\columnwidth]{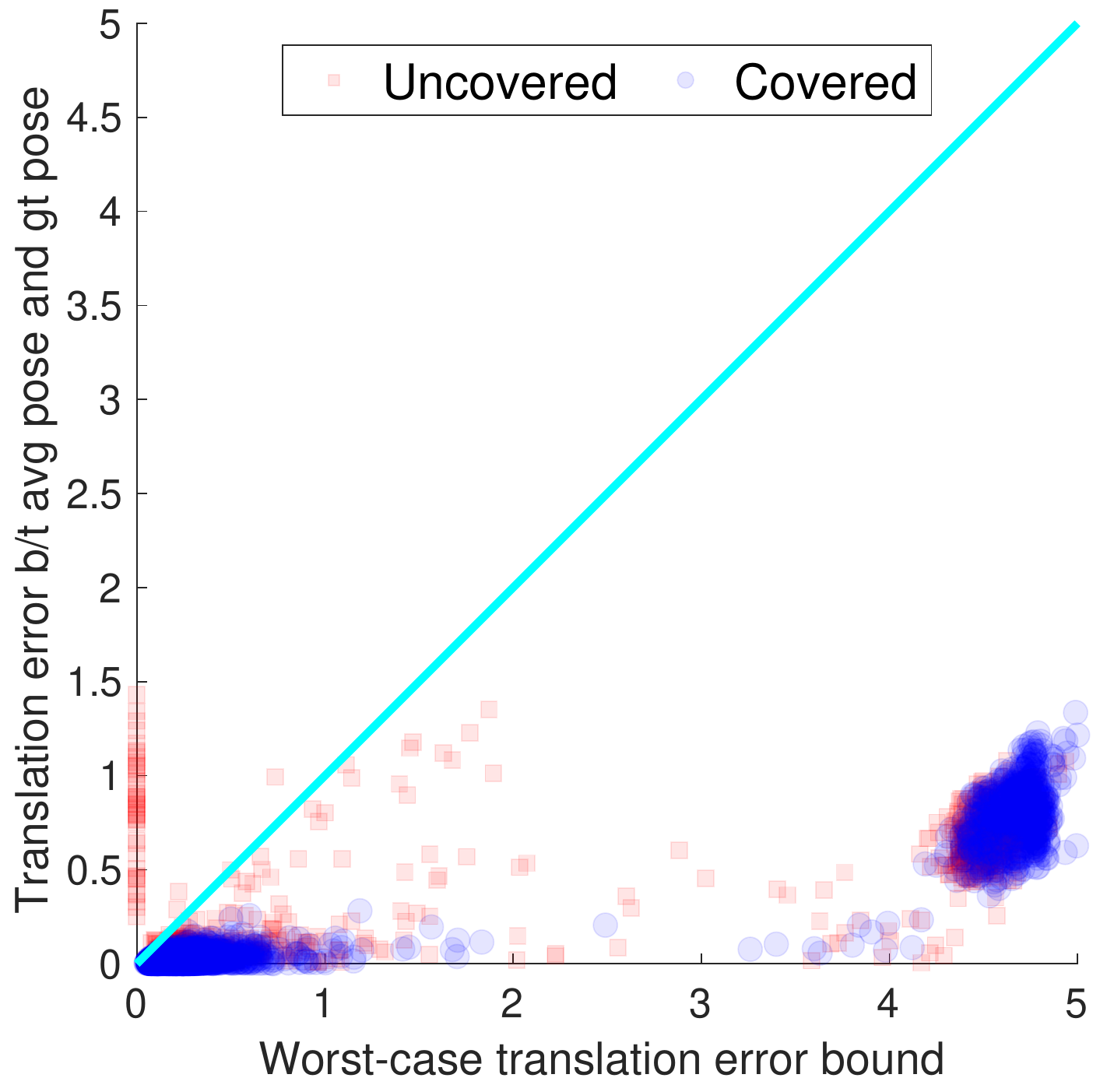}
		\end{minipage}
\end{tabular}
\vspace{-4mm}
\end{minipage}
\caption{Empirical coverage (left) and worst-case error bounds (middle: rotation, right: translation).  Top: $\epsilon=0.1$, bottom: $\epsilon=0.4$. For middle and right columns, $x$-axis represents the worst-case error bounds computed from~\eqref{eq:pose2purse}, $y$-axis represents the actual error between average pose and groundtruth pose. The area below the diagonal $y=x$ indicates correctness of the bounds (\ie, bound $\geq$ error), and points that are closer to the diagonal from below indicate \emph{tighter} bounds (perfect if precisely lie on the diagonal). Blue circles plot cases where the {\purse} covers the groundtruth pose and red squares plot cases were the {\purse} does not cover the groundtruth. Notice that blue circles never cross the diagonal and our bounds are correct when the \purse contains the pose (which holds with $1-\epsilon$ marginal probability).  
\label{fig:coverage-and-bound}} 
\end{center}
\vspace{-7mm}
\end{figure*}


\begin{table*}
\centering
\begin{adjustbox}{width=0.9\textwidth}
\begin{tabular}{c|cccc|cc|cc|cc|cc}
\hline
 & \multicolumn{4}{c}{Baselines (results adapted from~\cite{peng19cvpr-pvnet})} & \multicolumn{8}{|c}{Conformalized heatmap} \\
\cline{2-13}
  & Tekin & PoseCNN & Oberweger & PVNet & \multicolumn{2}{c|}{\gtball} & \multicolumn{2}{c|}{\gtellipse} & \multicolumn{2}{c|}{\frcnnball} & \multicolumn{2}{c}{\frcnnellipse}\\
 objects& \cite{tekin18cvpr-yolo} & \cite{xiang18rss-posecnn} & \cite{oberweger18eccv-heatmap} & \cite{peng19cvpr-pvnet} & $\epsilon=0.1$ & $\epsilon=0.4$ & $\epsilon=0.1$ & $\epsilon=0.4$ & $\epsilon=0.1$ & $\epsilon=0.4$ & $\epsilon=0.1$ & $\epsilon=0.4$ \\
\hline
ape & $7.01$ & $34.6$ & $69.6$ & $69.14$ & $77.70$ & $79.52$ & $79.26$  & $79.88$  & $70.20$ & $71.01$ & $68.84$ & $69.11$\\
can & $11.20$ & $15.10$ & $82.60$ & $86.09$ & $73.41$ & $75.97$ & $75.81$ & $78.13$ & $67.52$ & $69.81$ & $67.69$ & $69.56$ \\
cat & $3.62$ & $10.40$ & $65.10$ & $65.12$ & $87.36$ & $90.59$ & $89.54$  & $90.11$ & $74.95$ & $80.23$ & $68.98$ & $78.57$ \\
duck & $5.07$ & $31.80$ & $61.40$ & $61.44$ & $82.71$ & $83.08$ & $84.02$  & $83.55$ & $79.30$ & $80.62$ & $80.06$ & $80.53$ \\
driller & $1.40$ & $7.40$ & $73.80$ & $73.06$ & $79.32$ & $82.54$ & $81.22$  & $82.04$  & $58.48$ & $65.92$ & $58.06$ & $65.67$ \\
eggbox & - & $1.90$ & $13.10$ & $8.43$ & $0$ & $0$ & $0.09$ & $0.18$  & $0$ & $0$ & $0$ & $0.14$ \\
glue & $4.70$ & $13.80$ & $54.90$ & $55.37$ & $56.49$ & $71.08$ & $71.69$ & $72.93$ & $30.03$ & $47.18$ & $41.96$ & $48.26$ \\
holepuncher & $8.26$ & $23.10$ & $66.40$ & $69.84$ & $81.65$ & $82.89$ & $83.22$ & $84.30$ & $74.96$ & $77.85$ & $76.28$ & $78.18$ \\
\hline
average & $6.16$ & $17.20$ & $60.90$ & $61.06$ & $67.33$ & $70.71$ & $70.61$ & $71.39$ & $56.93$ & $61.58$ & $57.73$ & $61.25$\\
\hline
\end{tabular}
\end{adjustbox}
\vspace{-3mm}
\caption{Success rates of baseline methods and our conformalized heatmap (using the average pose) based on the 2D projection metric (\ie, a pose estimation is considered successful if the average 2D reprojection error is below $5$ pixels).\label{table:accuracy}}
\vspace{-6mm}
\end{table*}

\section{Experiments}
\label{sec:experiments}

We test our approach on the LineMOD Occlusion (\lmo) dataset~\cite{brachmann14eccv-linemodocc} to (i) justify the exchangeability assumption (Theorem~\ref{thm:icp-validity}) and suggest best practices for applying conformal prediction; (ii) evaluate the empirical coverage of the {\purse} and verify the correctness of Theorem~\ref{thm:icp-validity}, and (iii) compute the worst-case error bounds and demonstrate tightness or looseness. We also (iv) show that the average pose achieves better or similar accuracy as other approaches.

{\bf Implementation and runtime}. We set $T=1000$ in {\ransag}; use OpenGV~\cite{kneip14icra-opengv} for \pthreep and \pnp; and add a redundant $\norm{t} \leq 5$ in~\eqref{eq:purse} to ensure bounded translation. All procedures are implemented in Python except SDP relaxations are implemented in Matlab. The runtime of {\ransag} is comparable to {\ransac} and below one second. The runtime of computing~\eqref{eq:pose2purse} via SDPs is around $8$ seconds on a workstation with 2.2GHz AMD CPUs. The (second-order) SDP relaxations are almost always exact. 

{\bf Dataset and exchangeability}. The \lmo dataset contains $1214$ test images capturing $8$ different objects on a table, of which $200$ images were chosen by BOP19'20~\cite{hodan18eccv-bop}. We use the $200$ images for calibration and the entire $1214$ images for testing. As mentioned in Section~\ref{sec:pre:icp}, if the dataset was collected as a single video sequence under natural motion (\eg, a straight line), then the exchangeability assumption would fail. However, \cite{hinterstoisser12accv-linemod} described the data collection:
\begin{quote}
\vspace{-2mm}  
In order to guarantee a well distributed pose space sampling of the dataset pictures, we \emph{uniformly} divided the upper hemisphere of the objects into \emph{equally distant} pieces and took \emph{at most one image per piece}. As a result, our sequences provide \emph{uniformly distributed views} ...
\end{quote}
\vspace{-2mm}  
which indicates the $1214$ images are independent (\cf~\cite[Figs.~5-6]{hinterstoisser12accv-linemod}) and therefore exchangeable. This demonstrates a good example for data collection --to equally divide the parameter space and collect one observation per division-- so the guarantees offered by conformal prediction are valid.

{\bf Empirical coverage}. Our approach conformalizes the heatmaps~\cite{schmeckpeper22jfr-semantic} as~\eqref{eq:icp-ball} or~\eqref{eq:icp-ellipse}. The implementation\footnote{\url{https://github.com/yufu-wang/6D_Pose}} of~\cite{schmeckpeper22jfr-semantic} uses either groundtruth or Faster RCNN~\cite{ren15neurips-frcnn} bounding boxes, giving four variants of our approach: groundtruth box plus~\eqref{eq:icp-ball} or~\eqref{eq:icp-ellipse} (labels: \gtball, \gtellipse), and Faster RCNN box plus~\eqref{eq:icp-ball} or~\eqref{eq:icp-ellipse} (labels: \frcnnball, \frcnnellipse). Fig.~\ref{fig:coverage-and-bound} left column shows the empirical coverage (\ie, the percentage of images whose groundtruth poses lie in~\eqref{eq:purse}) of all four variants with $\epsilon=0.1$ and $\epsilon=0.4$. We see the empirical coverage is around $90\%$ when $\epsilon=0.1$ and around $60\%$ when $\epsilon=0.4$, for all $8$ objects. Though the empirical coverage can deviate from $1-\epsilon$, it generally stays within $\pm 5\%$ and mostly goes above $1-\epsilon$, which is encouraging given that our calibration set only has size $n=200$. Fig.~\ref{fig:methodoverview} (b) plots examples of the prediction sets. More examples are shown in the Supplementary Material.

{\bf Worst-case error bounds}. Fig.~\ref{fig:coverage-and-bound} middle column plots the worst-case rotation error bound ($x$-axis) vs. the actual rotation error between the average pose and the groundtruth ($y$-axis) for our approach using the {\gtball} setup (results for {\gtellipse}, {\frcnnball} and {\frcnnellipse} are similar and provided in the Supplementary Material). First, when the {\purse} covers the groundtruth (blue circles), the rotation error bound is always larger than the actual error (\ie, the blue circles never cross the $y=x$ diagonal). Second, when the error rate is increased from $\epsilon=0.1$ to $\epsilon=0.4$, we observe a shift of the blue circles towards $y=x$, indicating the error bounds get tightened. Third, our bounds are reasonably tight for most test images (\ie, the bottom-left cluster of blue circles) especially when $\epsilon=0.4$. However, they can become overly conservative (\ie, the line of blue circles on the right-side boundary) due to the keypoint prediction sets become too large. Fig.~\ref{fig:coverage-and-bound} right column plots similar results for the translation. The Supplementary Material gives a more detailed analysis of this conservatism, wherein we also solve~\eqref{eq:pose2purse} for multiple samples computed by {\ransag}, choose the minimum bound, and compare them with those obtained for the average pose (\cf Remark~\ref{rmk:bestbound}). 

{\bf Accuracy of the average pose}. We compare the accuracy of our average pose with other methods according to the 2D projection metric (an estimation is correct if the mean reprojection error is below $5$ pixels). Table~\ref{table:accuracy} shows: (i) our average pose achieves significantly better success rates when using groundtruth bounding boxes, and similar success rates when using Faster RCNN; (ii) the accuracy of the average pose increases when $\epsilon$ increases. 
\vspace{-2mm}
\section{Conclusions}
\label{sec:conclusion}
\vspace{-1mm}
We applied inductive conformal prediction to conformalize heatmap predictions as circular or elliptical prediction sets that guarantee probabilistic coverage of the groundtruth keypoints, propagated the uncertainty in keypoints to the object pose to form a {\purse}, designed {\ransag} to sample from {\purse} and compute an average pose, and used SDP relaxations to bound worst-case estimation errors. We validated our theory on the LineMOD Occlusion dataset. Future research will investigate better nonconformity functions,
 and applications to other vision problems.

\section*{Acknowledgement}
We thank Yufu Wang and Kostas Daniilidis for providing the heatmap implementation to detect semantic keypoints; Rachel Luo for discussing the exchangeability assumption in conformal prediction; Luca Carlone for pointing out related work on unknown-but-bounded noise (set membership) estimation in control theory, and anonymous reviewers for providing valuable feedback.

\clearpage
\onecolumn

\begin{center}
    {\Large \bf Supplementary Material}
\end{center}

\setcounter{section}{0}
\renewcommand{\thesection}{A\arabic{section}}
\renewcommand{\theequation}{A\arabic{equation}}
\renewcommand{\thetheorem}{A\arabic{theorem}}
\renewcommand{\thefigure}{A\arabic{figure}}
\renewcommand{\thetable}{A\arabic{table}}


\section{Proof of Proposition~\ref{prop:invariance}}

\begin{proof} Let $\{ \alpha_{i}  \}_{i=1}^n$ be the calibration scores obtained by applying $r$ in \eqref{eq:maxnonconformity} to the calibration set, and $\{ \alpha^{\rho}_i \}_{i=1}^{n}$ be the scores obtained by applying $r_{\rho}$ in \eqref{eq:robustnonconformity}. Observe that $\alpha^{\rho}_i = \rho(\alpha_i)$ because $\rho$ being monotonically increasing implies $\max \circ \rho = \rho \circ \max$ (``$\circ$'' describes function composition). As a result, it follows that $\alpha^{\rho}_{\pi(\floor{(n+1)\epsilon})} = \rho(\alpha_{\pi(\floor{(n+1)\epsilon})})$.
Let $\Feps_{\rho}$ be the ICP set due to $r_{\rho}$ for a given $\epsilon$, we have
\bea
\Feps_{\rho} = \{ \vy \in \calY \mid  \max\{ \rho(\phi(y_k,f(x)_k)) \}_{k=1}^K \leq \alpha^{\rho}_{\pi(\floor{(n+1)\epsilon})}  \} \nonumber \\
= \{ \vy \in \calY \mid  \rho( \max\{ \phi(y_k,f(x)_k) \}_{k=1}^K ) \leq \rho(\alpha_{\pi(\floor{(n+1)\epsilon})} ) \} \nonumber\\
= \{ \vy \in \calY \mid \max\{ \phi(y_k,f(x)_k) \}_{k=1}^K \leq \alpha_{\pi(\floor{(n+1)\epsilon})} \}, \nonumber 
\eea
where the last set is precisely $\Feps$, the ICP set induced by $r$.
\end{proof}


\section{Proof of Proposition~\ref{prop:purse}}
\begin{proof} Recall the ICP set in~\eqref{eq:icpunify}
\bea\label{eq:icpunifyrestate}
\Feps(x) = \cbrace{\vy \in \calY \mid (y_k - \mu_k)\tran \Lambda_k (y_k - \mu_k) \leq 1,\forall k}
\eea
that defines either a \eqref{eq:icp-ball} or an \eqref{eq:icp-ellipse}. From the pinhole camera projection model, we know that the groundtruth keypoints $\vy = (y_1,\dots,y_K)$ satisfy
\bea\label{eq:projection}
y_k = \Pi (\Rgt Y_k + \tgt) = \frac{[P(\Rgt Y_k + \tgt)]_{1:2}}{[P(\Rgt Y_k + \tgt)]_{3}}, k=1,\dots,K 
\eea
where $P \in \Real{3\times 3}$ denotes the camera intrinsics, $Y_k \in \Real{3}$ is location of the $k$-th 3D keypoint in the object's coordinate frame, $[\vv]_{1:2}$ (resp. $[\vv]_3$) denotes the first two (resp. third) entries of a 3D vector $\vv$. To simplify our notation, we develop~\eqref{eq:projection} as
\bea
P\Rgt Y_k + P \tgt = (Y_k\tran \kron P) \vectorize{\Rgt} + P \tgt 
= \underbrace{\bmat{cc} Y_k\tran \kron P & P \emat}_{:= U_{k} \in \Real{3 \times 12}}
\bmat{c} \vectorize{\Rgt} \\ \tgt \emat = \bmat{c} u_{k,1}\tran \\ u_{k,2}\tran \\ u_{k,3}\tran \emat \sgt \\
\Longrightarrow y_k = \parentheses{\bmat{c} u_{k,1}\tran \\ u_{k,2}\tran \emat \sgt} \bigg/ (u_{k,3}\tran \sgt), \label{eq:simpleyk}
\eea
where $u_{k,j}\tran \in \Real{1 \times 12}$ denotes the $j$-th row of matrix $U_k$. Notice that $u_{k,3}\tran \sgt$ is the depth of the $k$-th 3D keypoint in the camera coordinate frame (after rigid transformation $(\Rgt,\tgt)$).

{\bf In front of the camera}. Since the camera observes the object, the groundtruth pose $\sgt$ must transform the object to lie in front of the camera. Therefore, the keypoints must have positive depth values:
\bea
u_{k,3}\tran \sgt > 0, k=1,\dots,K.
\eea

{\bf Within ICP sets}. We now insert~\eqref{eq:simpleyk} back to the constraint defined by the ICP set~\eqref{eq:icpunifyrestate}, leading to
\bea
(y_k - \mu_k)\tran \Lambda_k (y_k - \mu_k) \leq 1 \Longleftrightarrow \\ 
\frac{1}{(u_{k,3}\tran \sgt)^2} \sgt\tran \bmat{cc} u_{k,1} - \mu_{k,1} u_{k,3} & u_{k,2} - \mu_{k,2} u_{k,3} \emat \Lambda_k  \bmat{c} u_{k,1}\tran - \mu_{k,1} u_{k,3}\tran \\ u_{k,2}\tran - \mu_{k,2} u_{k,3} \emat \sgt \leq 1 \Longleftrightarrow \\
\sgt\tran \bmat{cc} u_{k,1} - \mu_{k,1} u_{k,3} & u_{k,2} - \mu_{k,2} u_{k,3} \emat \Lambda_k  \bmat{c} u_{k,1}\tran - \mu_{k,1} u_{k,3}\tran \\ u_{k,2}\tran - \mu_{k,2} u_{k,3} \emat \sgt \leq \sgt\tran \parentheses{u_{k,3} u_{k,3}\tran} \sgt \Longleftrightarrow \\
\sgt\tran \underbrace{\parentheses{ \bmat{cc} u_{k,1} - \mu_{k,1} u_{k,3} & u_{k,2} - \mu_{k,2} u_{k,3} \emat \Lambda_k  \bmat{c} u_{k,1}\tran - \mu_{k,1} u_{k,3}\tran \\ u_{k,2}\tran - \mu_{k,2} u_{k,3} \emat - u_{k,3} u_{k,3}\tran }}_{: = A_k \in \sym{12}} \sgt \leq 0, \label{eq:Ak}
\eea
which indicates that the groundtruth pose $\sgt$ must satisfy $K$ quadratic constraints, one for each keypoint.
In summary, the groundtruth pose $\sgt$ must lie in the \eqref{eq:purse} with $b_k = u_{k,3}$ and $A_k$ as in~\eqref{eq:Ak}.
\end{proof}

\section{Supplementary Experiments}

\begin{figure}[h]
\begin{center}
\begin{minipage}{\textwidth}
\centering
\begin{tabular}{cc}%
	    \begin{minipage}{5cm}%
		\centering%
		\includegraphics[width=\columnwidth]{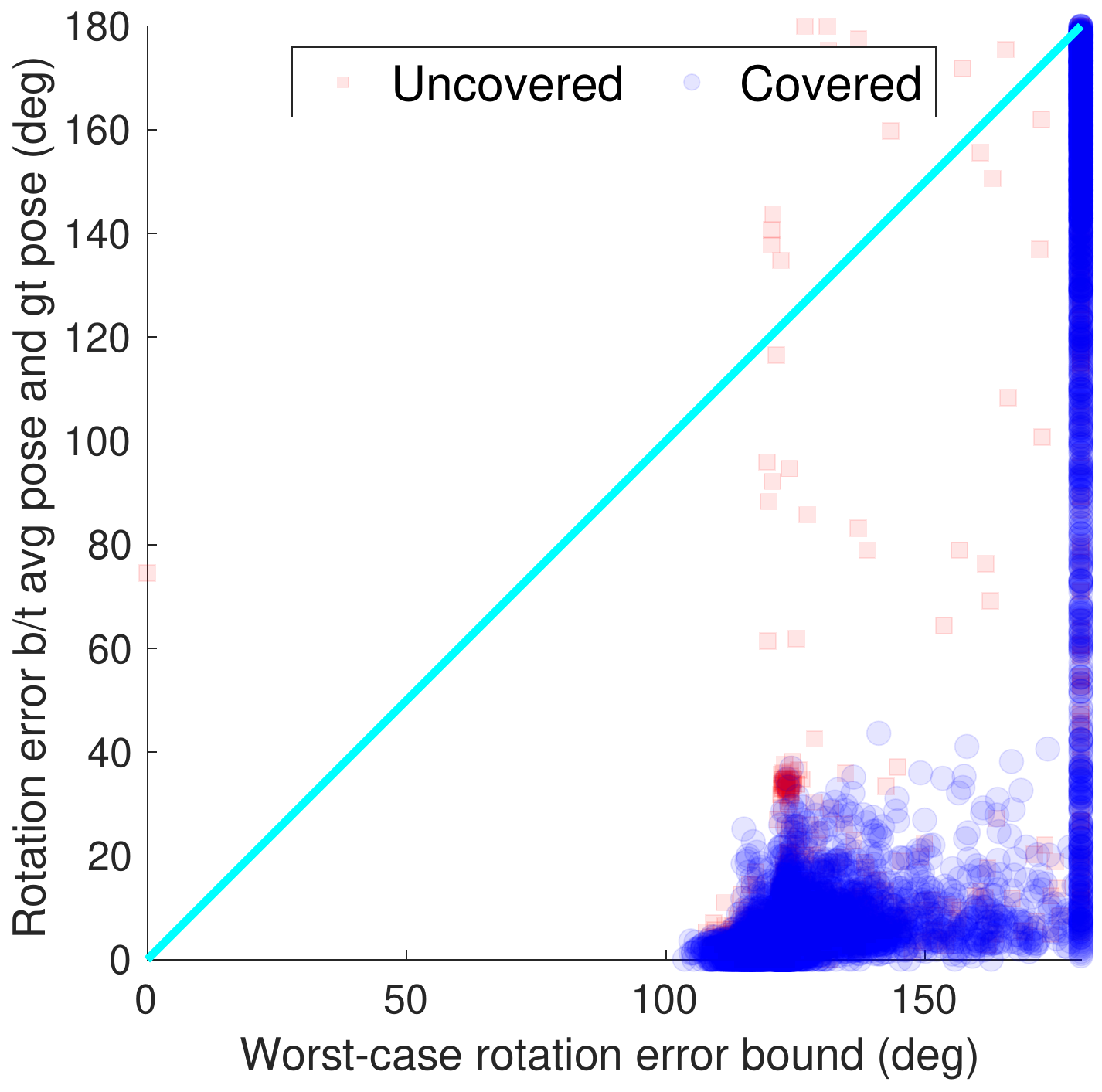}
		\end{minipage}
	&  	
	    \begin{minipage}{5cm}%
		\centering%
		\includegraphics[width=\columnwidth]{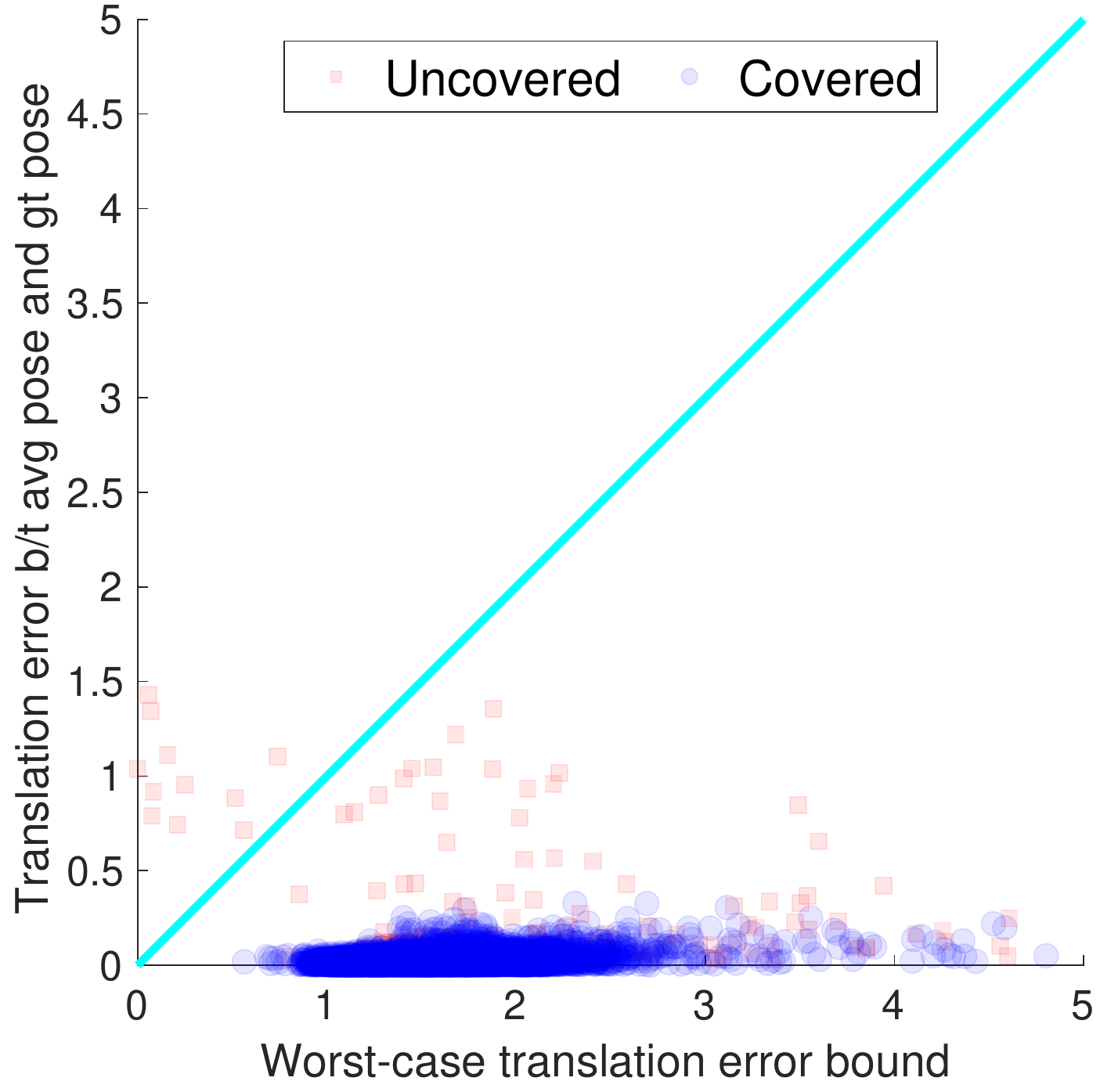}
		\end{minipage}
	\\
	\multicolumn{2}{c}{(a) $\epsilon=0.1$}
	\\
	    \begin{minipage}{5cm}%
		\centering%
		\includegraphics[width=\columnwidth]{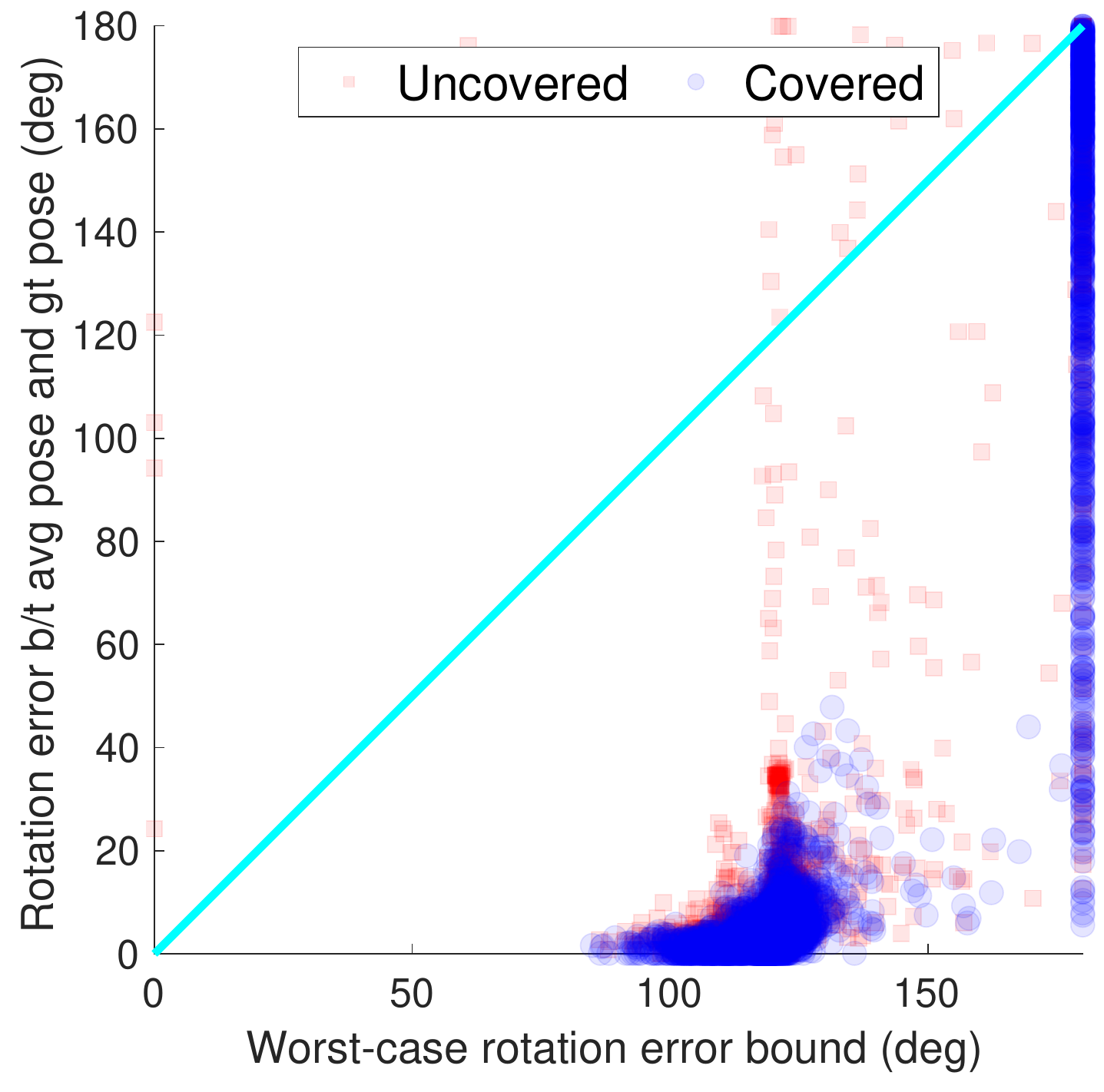}
		\end{minipage}
	&  
	    \begin{minipage}{5cm}%
		\centering%
		\includegraphics[width=\columnwidth]{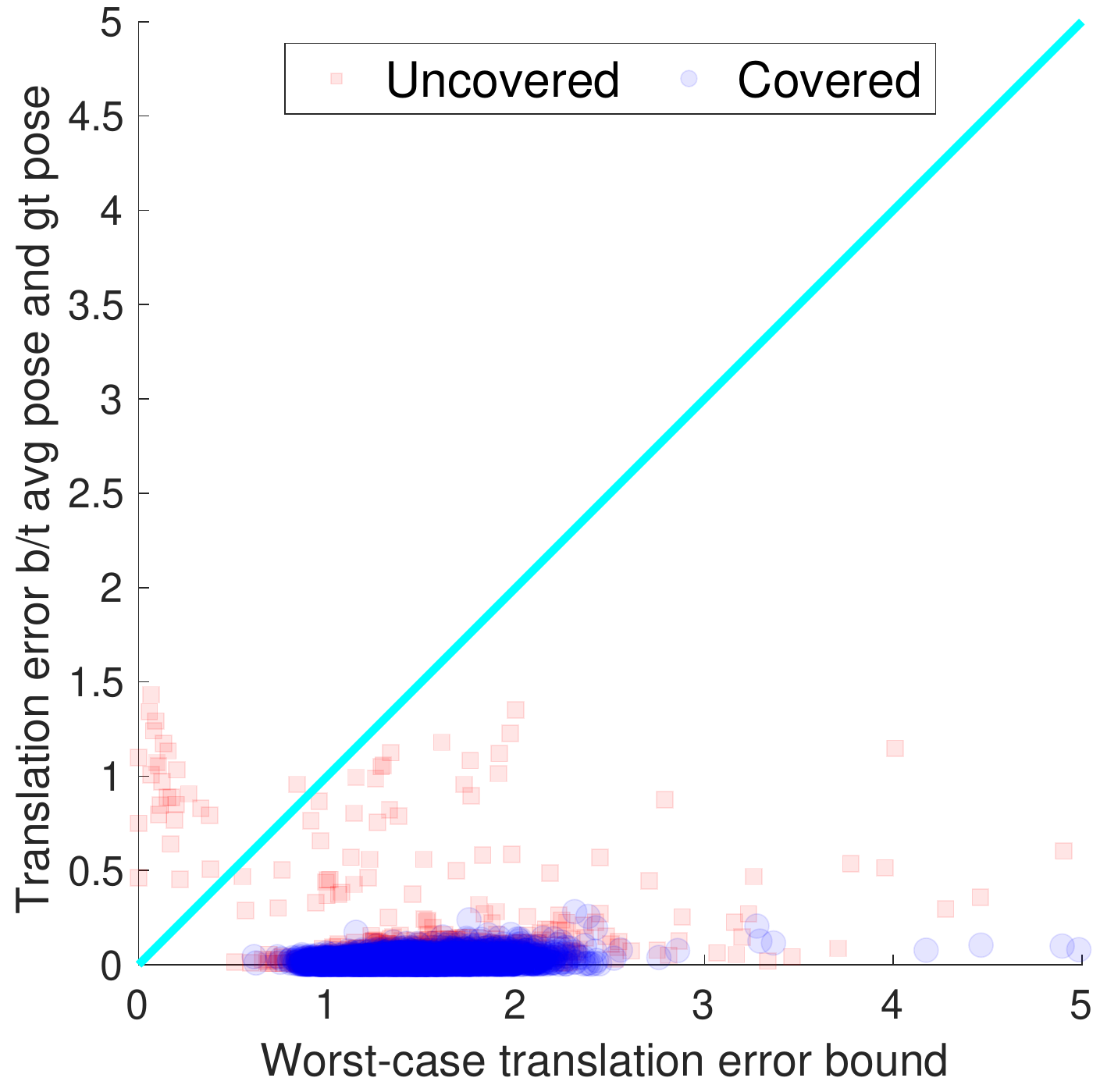}
		\end{minipage}
	\\
	\multicolumn{2}{c}{(b) $\epsilon=0.4$}
\end{tabular}
\end{minipage}
\caption{Looser (albeit faster) worst-case error bounds computed from solving first-order relaxation of~\eqref{eq:pose2purse}, compared to worst-case error bounds computed from solving second-order relaxation shown in Fig.~\ref{fig:coverage-and-bound} middle and right columns. \label{fig:relax-order}} 
\end{center}
\vspace{-7mm}
\end{figure}

\subsection{Ablation: Relaxation Order}
In the main document, we briefly described that we applied second-order semidefinite relaxations to compute the worst-case error bounds in~\eqref{eq:pose2purse} and reported that the average runtime is around $8$ seconds on an ordinary workstation. Here we justify the choice of second-order relaxations by showing that first-order relaxations, although much faster (average runtime is about $0.1$ seconds), lead to much looser upper bounds for the optimization~\eqref{eq:pose2purse}. 

To help the reader better understand the approach, we first give a very short introduction to semidefinite relaxations for polynomial optimization problems (POPs). We refer the reader to~\cite[Section 2]{yang22pami-certifiably} for a detailed introduction.  

{\bf Polynomial optimization problems} (POPs) are problems of the following general formulation
\bea
\min_{x \in \Real{n}} & p(x) \\
\subject & h_i(x) = 0, i=1,\dots,l_h,\\
& g_j(x) \geq 0, j=1,\dots,l_g
\eea 
where $p,\{h_i\}_{i=1}^{l_h}, \{ g_j\}_{j=1}^{l_g}$ are all polynomial functions in $x \in \Real{n}$. Notice that if we denote $s = [\vectorize{R}\tran,t\tran]\tran \in \Real{12}$, it is clear that the cost function of~\eqref{eq:pose2purse} is a polynomial in $s$ when fixing a particular $\lambda$ (we can add a minus sign to the cost of~\eqref{eq:pose2purse} so that we convert ``$\max$'' to ``$\min$''). The constraints for~\eqref{eq:pose2purse} is $(R,t) \in \Seps$ where $\Seps$ has the form in~\eqref{eq:purse}. We claim that the~\eqref{eq:purse} can be described by a set of polynomial equalities and inequalities. This is because (i) the rotation constraint $R \in \SOthree$ can be described by $15$ quadratic equality constraints~\cite{yang22pami-certifiably}; (ii) the quadratic constraints in~\eqref{eq:purse} are already polynomial constraints; and (iii) the linear inequalities $b_k\tran s > 0$ can be equivalently written as $b_k\tran s \geq \epsilon$ for a small $\epsilon > 0$ (note that $b_k\tran s$ is the depth of the 3D keypoints, so it makes sense to enforce they are larger than, say $\epsilon = 0.001$). We conclude that computing the worst-case error bounds~\eqref{eq:pose2purse} is a POP.

{\bf Semidefinite relaxations} are a powerful tool to approximate (or even exactly compute) the \emph{global optimal} solutions for (generally nonconvex) POPs. In particular, Lasserre's hierarchy of moment and sums-of-squares relaxations~\cite{lasserre01global} provides a systematic approach to design such semidefinite relaxations. In particular, Lasserre's hierarchy relaxes a POP into a hierarchy of convex semidefinite programs (SDPs) of increasing size. Each relaxation, at a so-called \emph{relaxation order}, in this hierarchy can be solved in polynomial time and provides a valid lower bound for the POP (if the POP aims to maximize, as in~\eqref{eq:pose2purse}, then a valid upper bound is provided). Moreover, under mild technical conditions, the lower (upper) bounds of these relaxations coincide with the global optimum of the original POP, in which case we say the relaxation is \emph{exact}, or \emph{tight}.

{\bf First-order vs. second-order relaxations}. The minimum relaxation order for the POP~\eqref{eq:pose2purse} is $1$, since all the polynomials in~\eqref{eq:pose2purse} have degree at most $2$ (in general, the minimum relaxation order for a POP is $\lceil d/2 \rceil$, where $d$ is the maximum degree of the polynomials defining a POP). In practice we choose a second-order relaxation instead of a first-order relaxation because first-order relaxations give loose upper bounds for problem~\eqref{eq:pose2purse}. Fig.~\ref{fig:relax-order} plots the worst-case error bounds computed by solving the first-order relaxation of~\eqref{eq:pose2purse} under the same {\gtball} setup. Compared to Fig.~\ref{fig:coverage-and-bound} middle and right columns, we clearly see that solving the first-order relaxation produces overly conservation upper bounds for~\eqref{eq:pose2purse}. For example, when $\epsilon=0.1$, solving the first-order relaxation never produces a rotation error bound that is below $100^\circ$, while in Fig.~\ref{fig:coverage-and-bound} we see a cluster of blue circles near the bottom left corner indicating tight bounds.

One nice property of applying semidefinite relaxations is that we get a certificate of global optimality when the relaxation is indeed exact. Such certificates typically come in the form of a rank-one optimal SDP solution, or a relative suboptimality gap (\cf \cite[eq. (24)]{yang22pami-certifiably}), which indicates exactness of the relaxation when the value is numerically zero (loosely speaking, a relative suboptimality gap of $\epsilon\%$ means that the global optimum of the SDP is at most $\epsilon$ percentage away from the global optimum of the POP). When we solve second-order relaxations of problem~\eqref{eq:pose2purse} under the {\gtball} setup with $\lambda = 1$, we obtain a relative suboptimality gap that is below $10^{-3}$ (resp. $10^{-6}$) for $99.02\%$ (resp. $72.51\%$) of the $8784$ test problems, indicating that the second-order relaxation is sufficient to obtain (approximately) globally optimal solutions for problem~\eqref{eq:pose2purse}.

\subsection{Qualitative ICP Sets}

Fig.~\ref{fig:methodoverview}(b) shows circular and elliptical examples of the ICP sets. Fig.~\ref{fig:heatmap-qualitative} provides more examples of the ICP sets with $\epsilon=0.1$ and $\epsilon=0.4$. Notice how the ICP sets become smaller when $\epsilon$ increases.

\newcommand{\myimgwidth}{0.85\textwidth}
\begin{figure*}[h]
\centering
\vspace{-8mm}
\includegraphics[width=\myimgwidth]{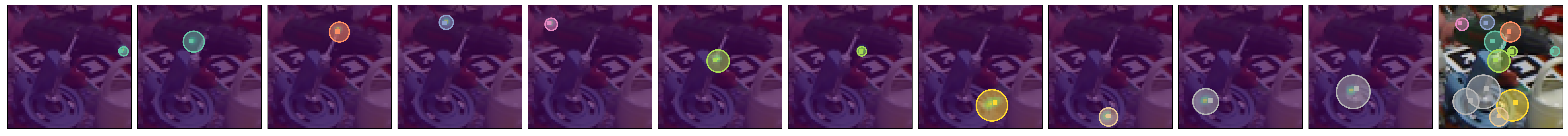}\\
\vspace{-1mm}
\includegraphics[width=\myimgwidth]{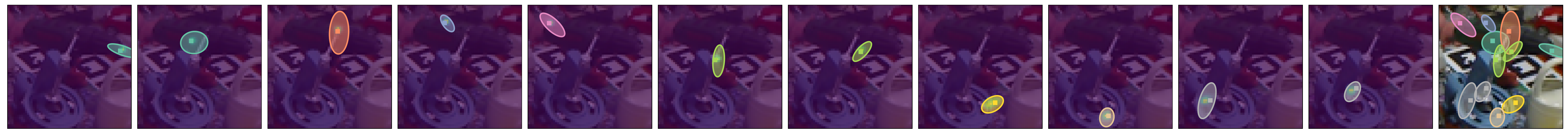}\\
\vspace{-1mm}
\includegraphics[width=\myimgwidth]{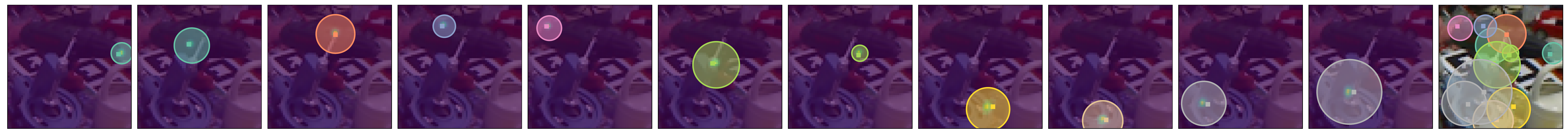}\\
\vspace{-1mm}
\includegraphics[width=\myimgwidth]{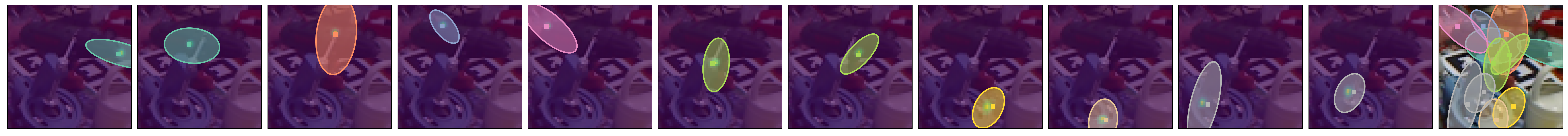}\\
\vspace{-2mm}
{\small (a) $\epsilon=0.1$, object: \emph{driller}, top to bottom: \gtball, \gtellipse, \frcnnball, \frcnnellipse  }\\
\includegraphics[width=\myimgwidth]{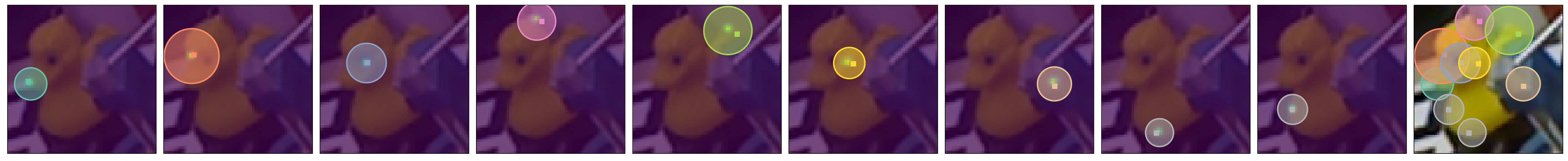}\\
\vspace{-1mm}
\includegraphics[width=\myimgwidth]{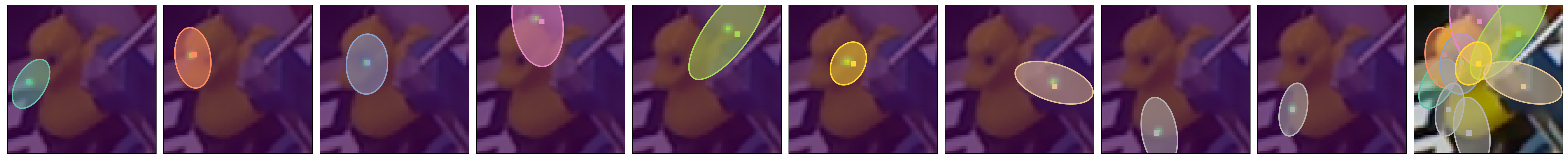}\\
\vspace{-1mm}
\includegraphics[width=\myimgwidth]{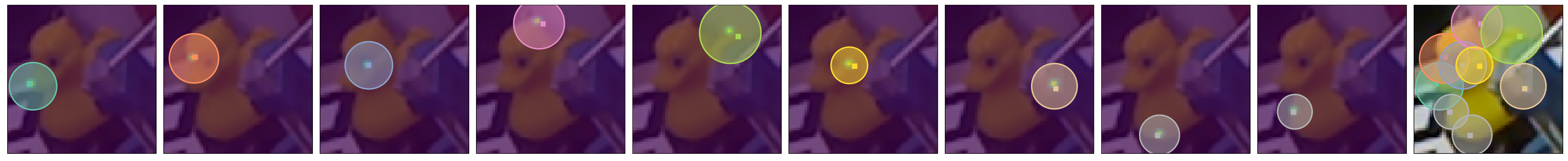}\\
\vspace{-1mm}
\includegraphics[width=\myimgwidth]{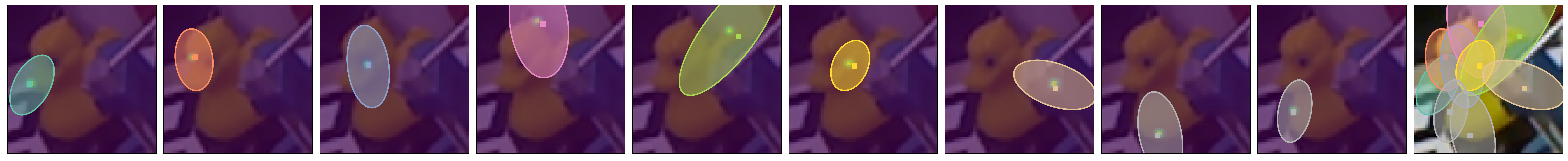}\\
\vspace{-2mm}
{\small (b) $\epsilon=0.1$, object: \emph{duck}, top to bottom: \gtball, \gtellipse, \frcnnball, \frcnnellipse  }\\
\includegraphics[width=\myimgwidth]{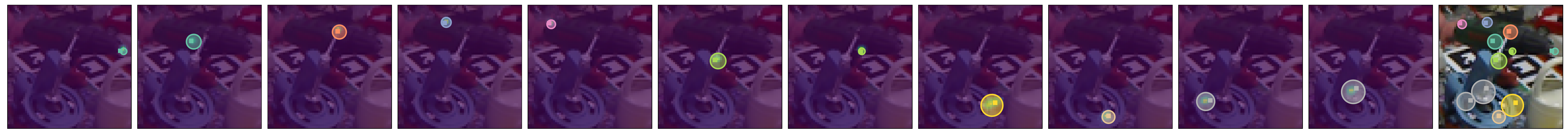}\\
\vspace{-1mm}
\includegraphics[width=\myimgwidth]{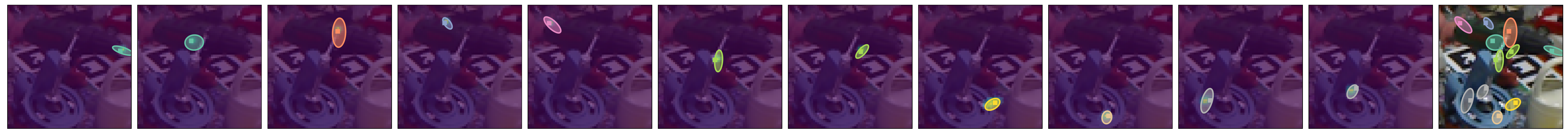}\\
\vspace{-1mm}
\includegraphics[width=\myimgwidth]{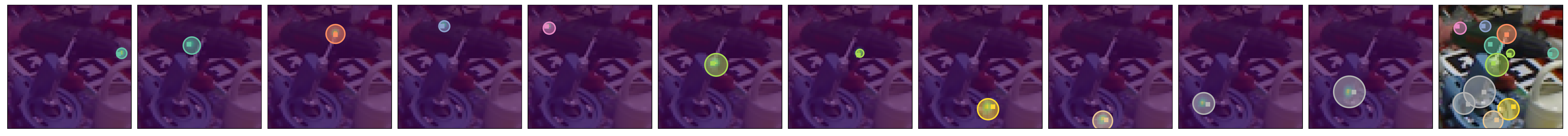}\\
\vspace{-1mm}
\includegraphics[width=\myimgwidth]{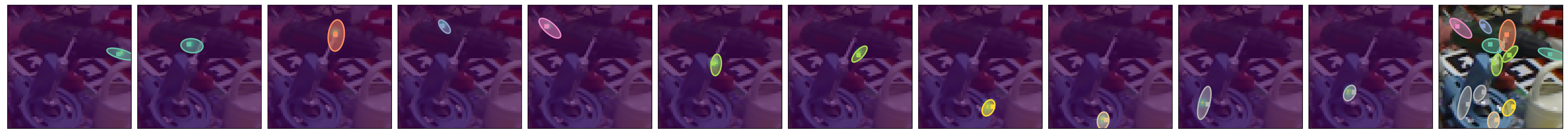}\\
\vspace{-2mm}
{\small (c) $\epsilon=0.4$, object: driller, top to bottom: \gtball, \gtellipse, \frcnnball, \frcnnellipse}\\
\includegraphics[width=\myimgwidth]{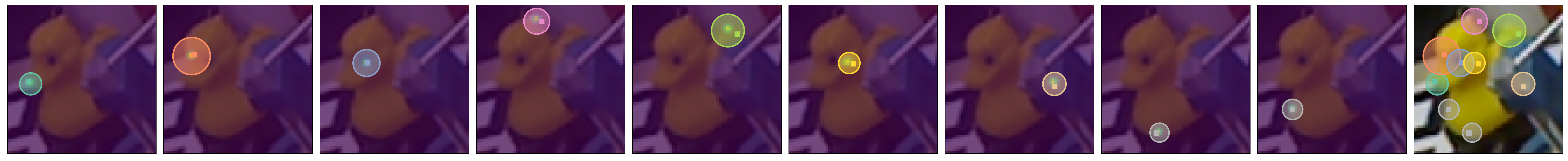}\\
\vspace{-1mm}
\includegraphics[width=\myimgwidth]{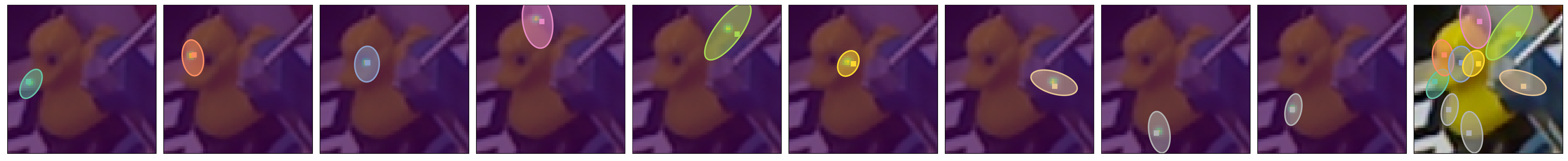}\\
\vspace{-1mm}
\includegraphics[width=\myimgwidth]{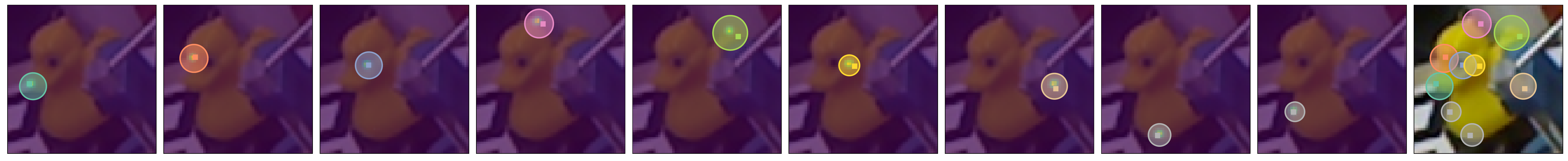}\\
\vspace{-1mm}
\includegraphics[width=\myimgwidth]{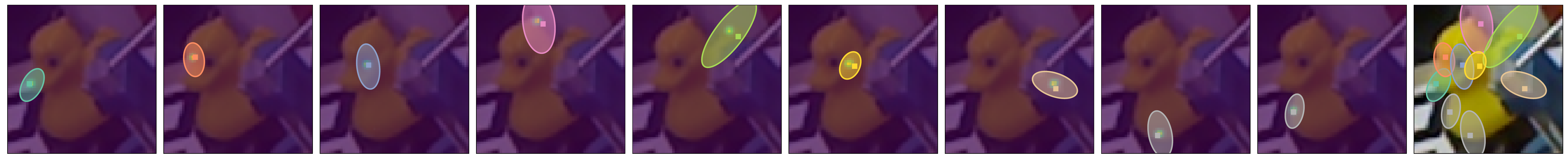}\\
\vspace{-2mm}
{\small (d) $\epsilon=0.4$, object: \emph{duck}, top to bottom: \gtball, \gtellipse, \frcnnball, \frcnnellipse  }\\
\vspace{-3mm}
\caption{ICP sets on {\lmo}~\cite{brachmann14eccv-linemodocc}. Last image of each row overlays \emph{all} groundtruth keypoints (squares) and ICP sets (balls \& ellipses) on the original image. Other images overlay the heatmap, groundtruth location, and ICP set of a \emph{single} keypoint on the original image.
\label{fig:heatmap-qualitative}} 
\end{figure*}

\subsection{Worst-case Error Bounds under {\gtellipse}, {\frcnnball}, and {\frcnnellipse} setups}
\begin{figure}[h]
\begin{center}
\begin{minipage}{\textwidth}
\centering
\begin{tabular}{cccc}%
	    \begin{minipage}{4cm}%
		\centering%
		\includegraphics[width=\columnwidth]{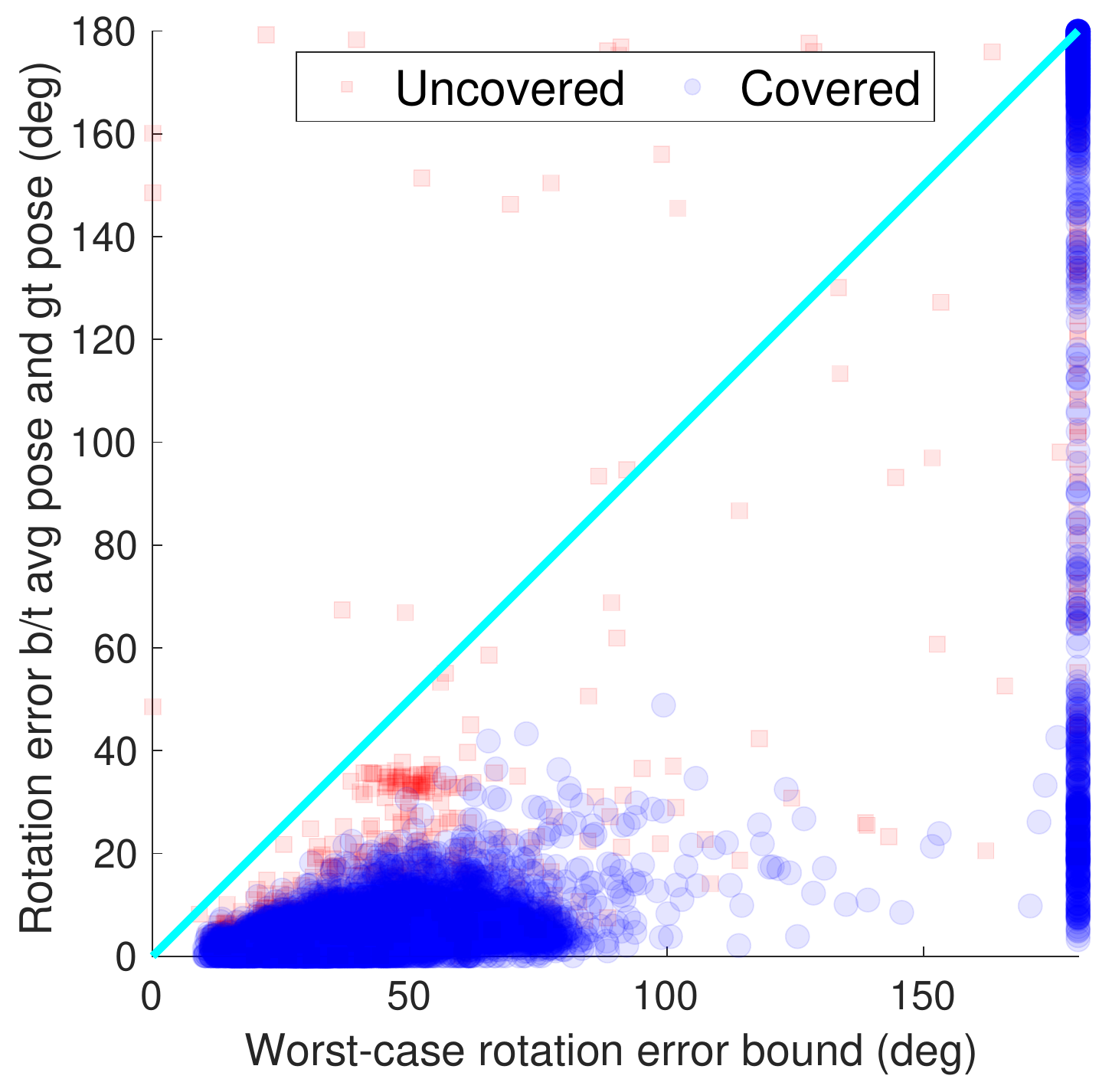}
		\end{minipage}
	&  	
	    \begin{minipage}{4cm}%
		\centering%
		\includegraphics[width=\columnwidth]{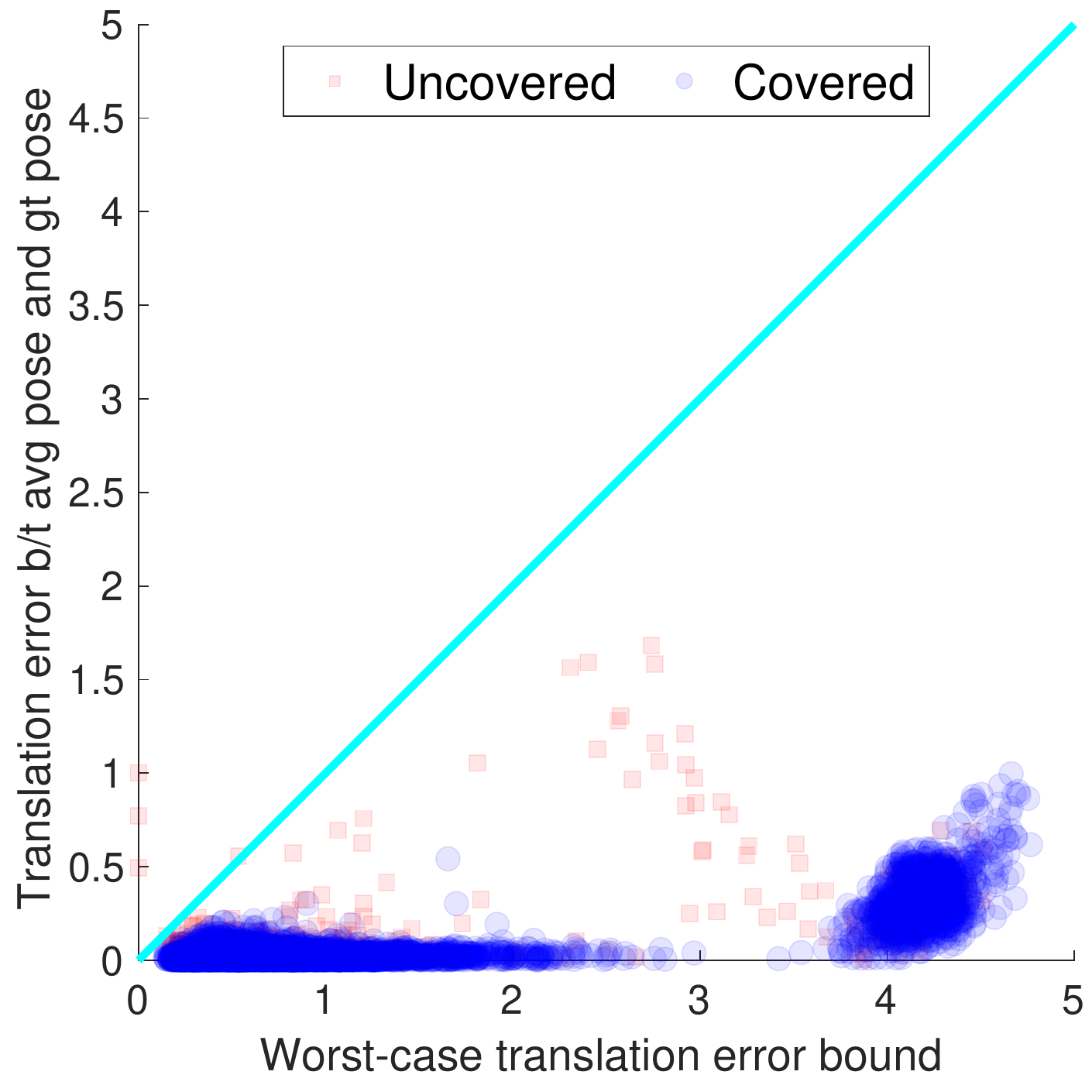}
		\end{minipage}
	&
	\begin{minipage}{4cm}%
		\centering%
		\includegraphics[width=\columnwidth]{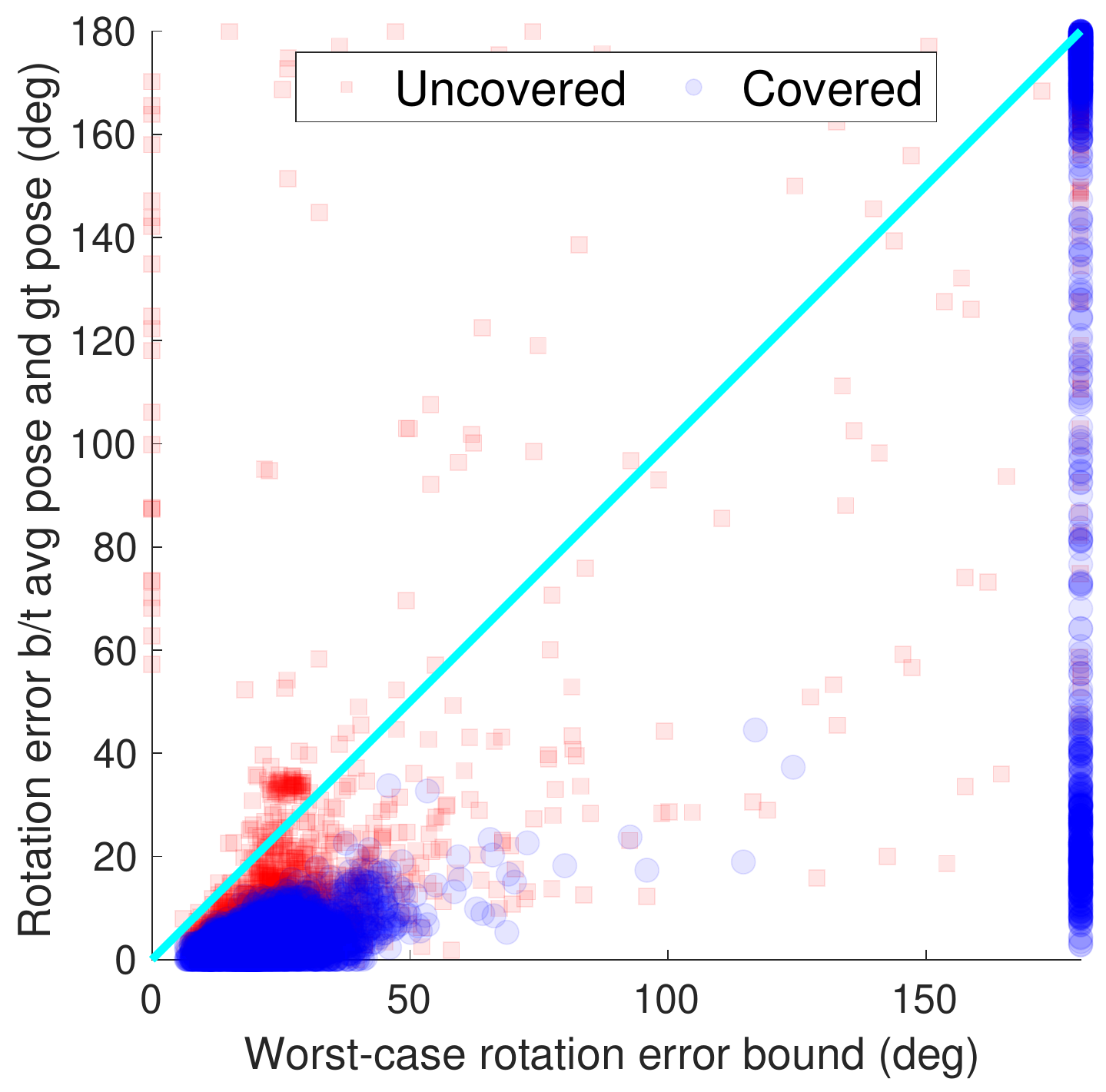}
		\end{minipage}
	&  	
	    \begin{minipage}{4cm}%
		\centering%
		\includegraphics[width=\columnwidth]{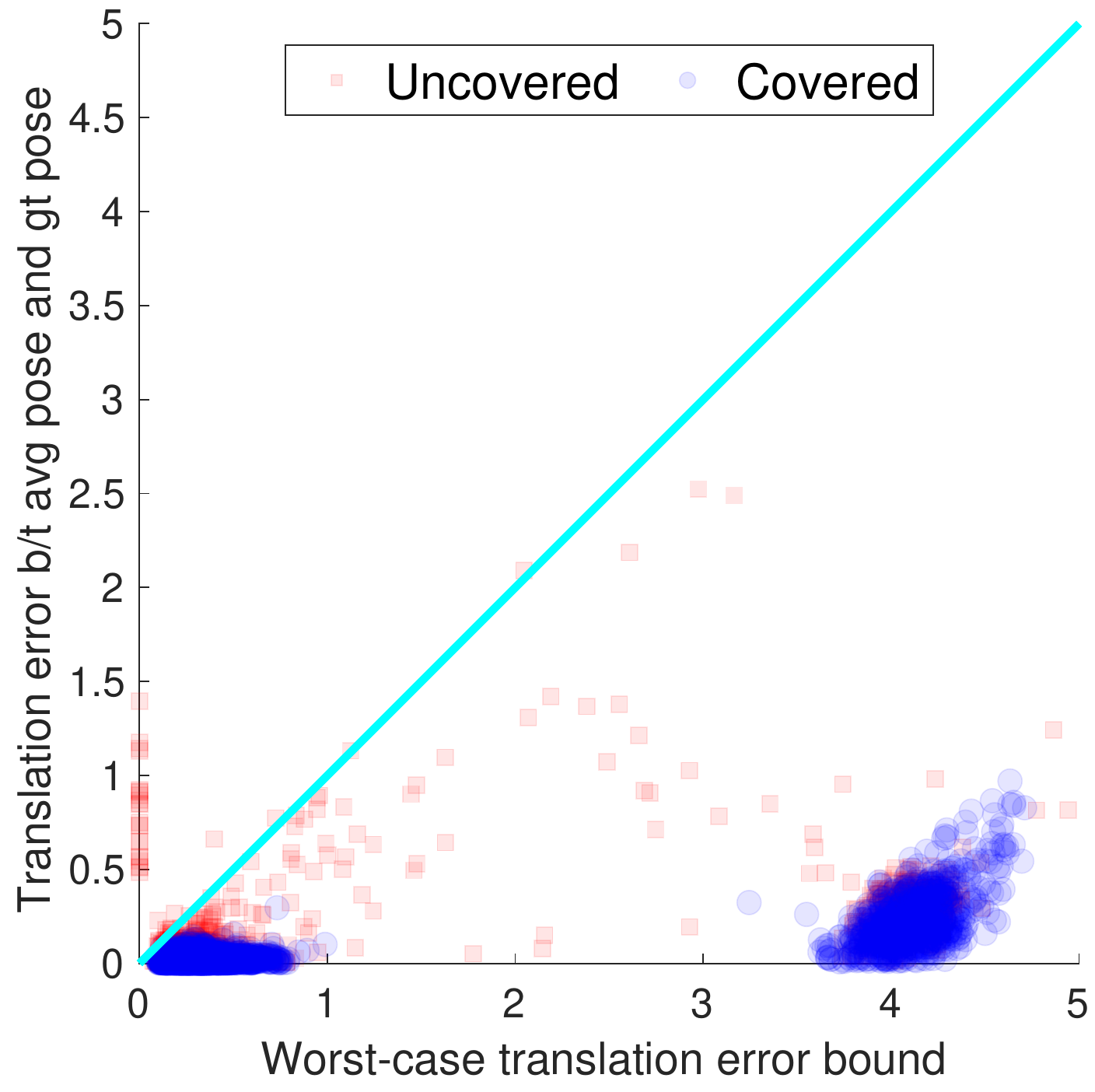}
		\end{minipage}
	\\
	\multicolumn{4}{c}{(a) \gtellipse. Left two columns: $\epsilon=0.1$; right two columns: $\epsilon=0.4$. }
	\\
	    \begin{minipage}{4cm}%
		\centering%
		\includegraphics[width=\columnwidth]{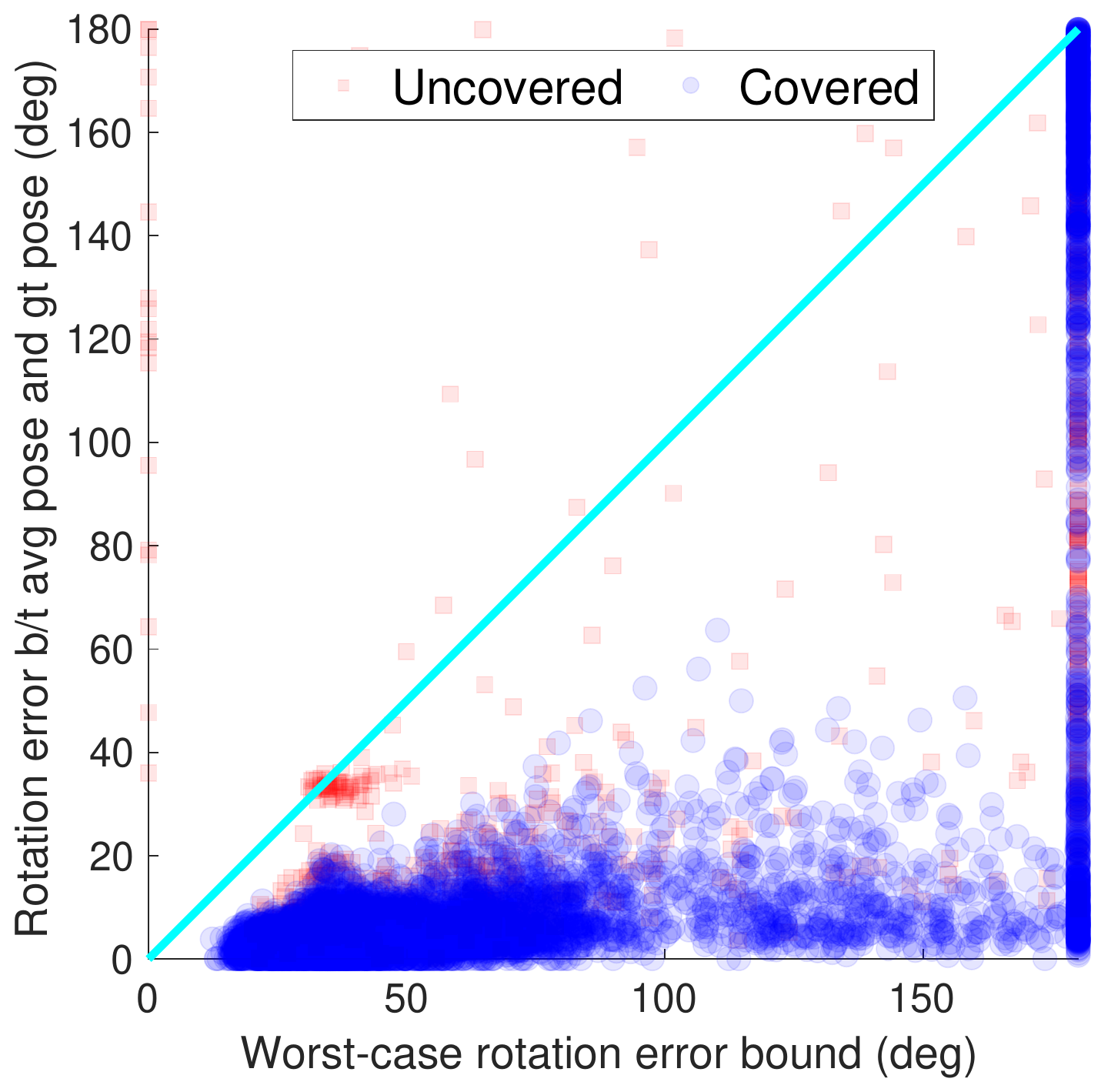}
		\end{minipage}
	&  	
	    \begin{minipage}{4cm}%
		\centering%
		\includegraphics[width=\columnwidth]{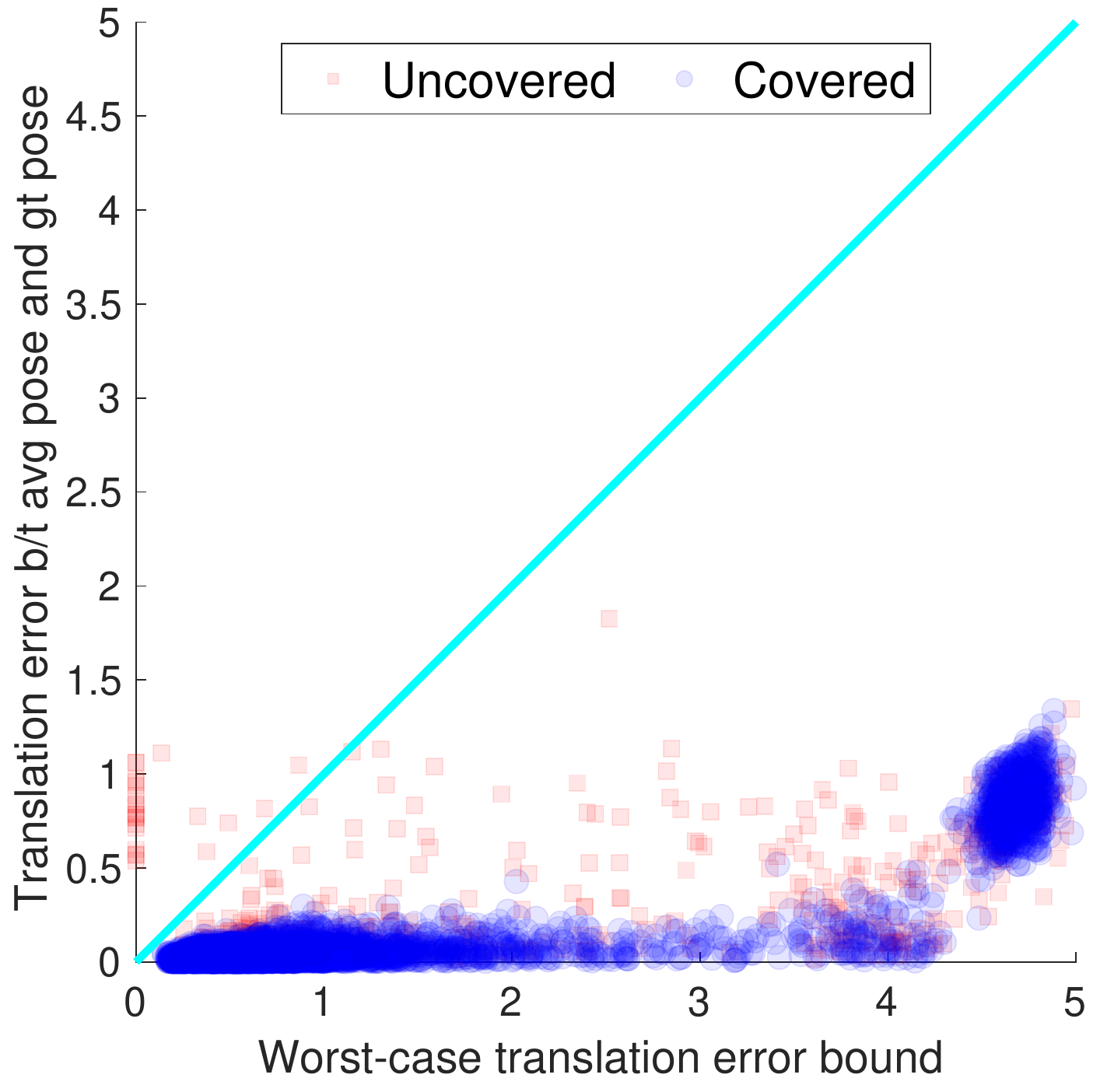}
		\end{minipage}
	&
	\begin{minipage}{4cm}%
		\centering%
		\includegraphics[width=\columnwidth]{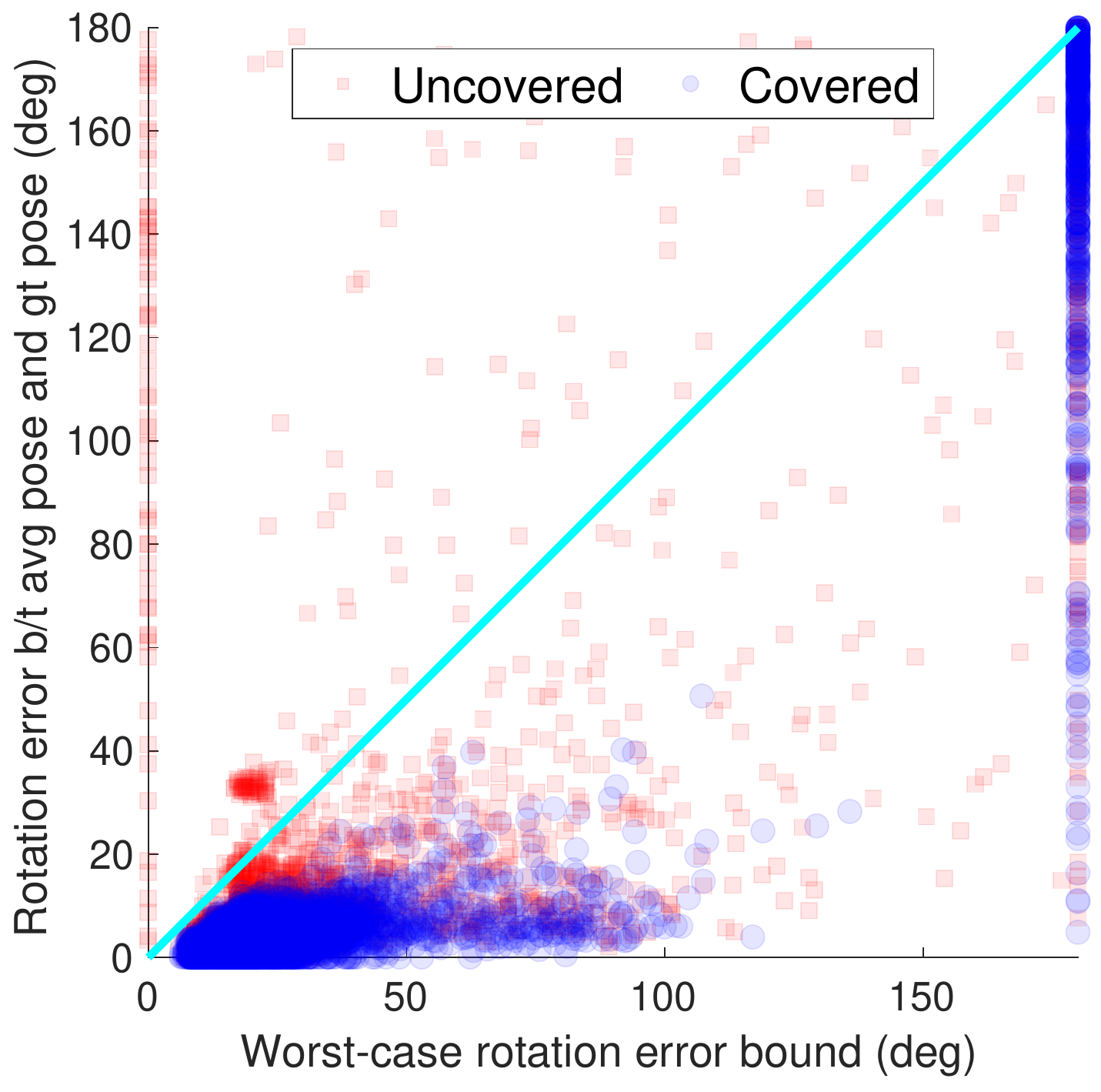}
		\end{minipage}
	&  	
	    \begin{minipage}{4cm}%
		\centering%
		\includegraphics[width=\columnwidth]{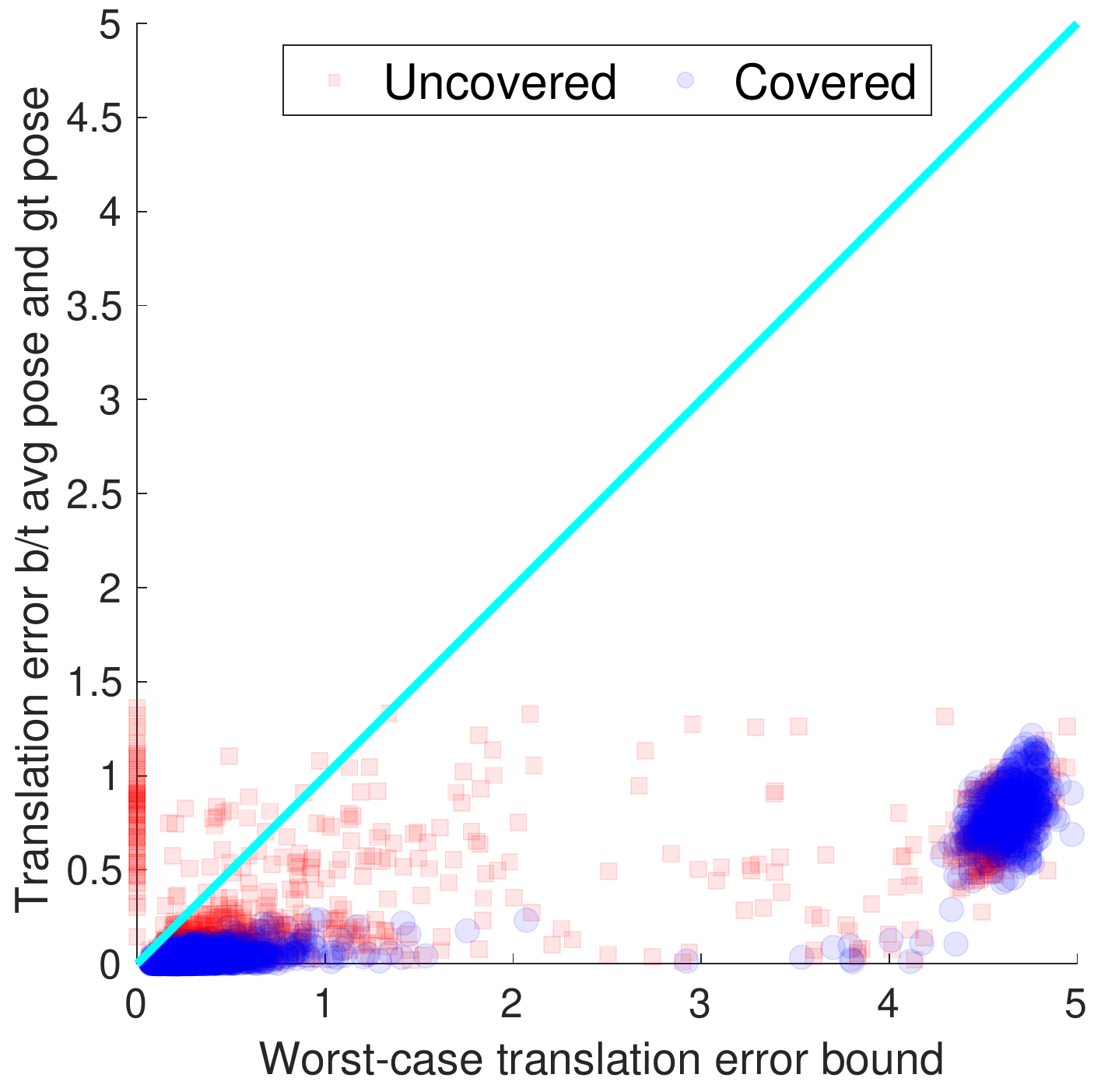}
		\end{minipage}
	\\
	\multicolumn{4}{c}{(b) \frcnnball. Left two columns: $\epsilon=0.1$; right two columns: $\epsilon=0.4$. }
	\\
	\begin{minipage}{4cm}%
		\centering%
		\includegraphics[width=\columnwidth]{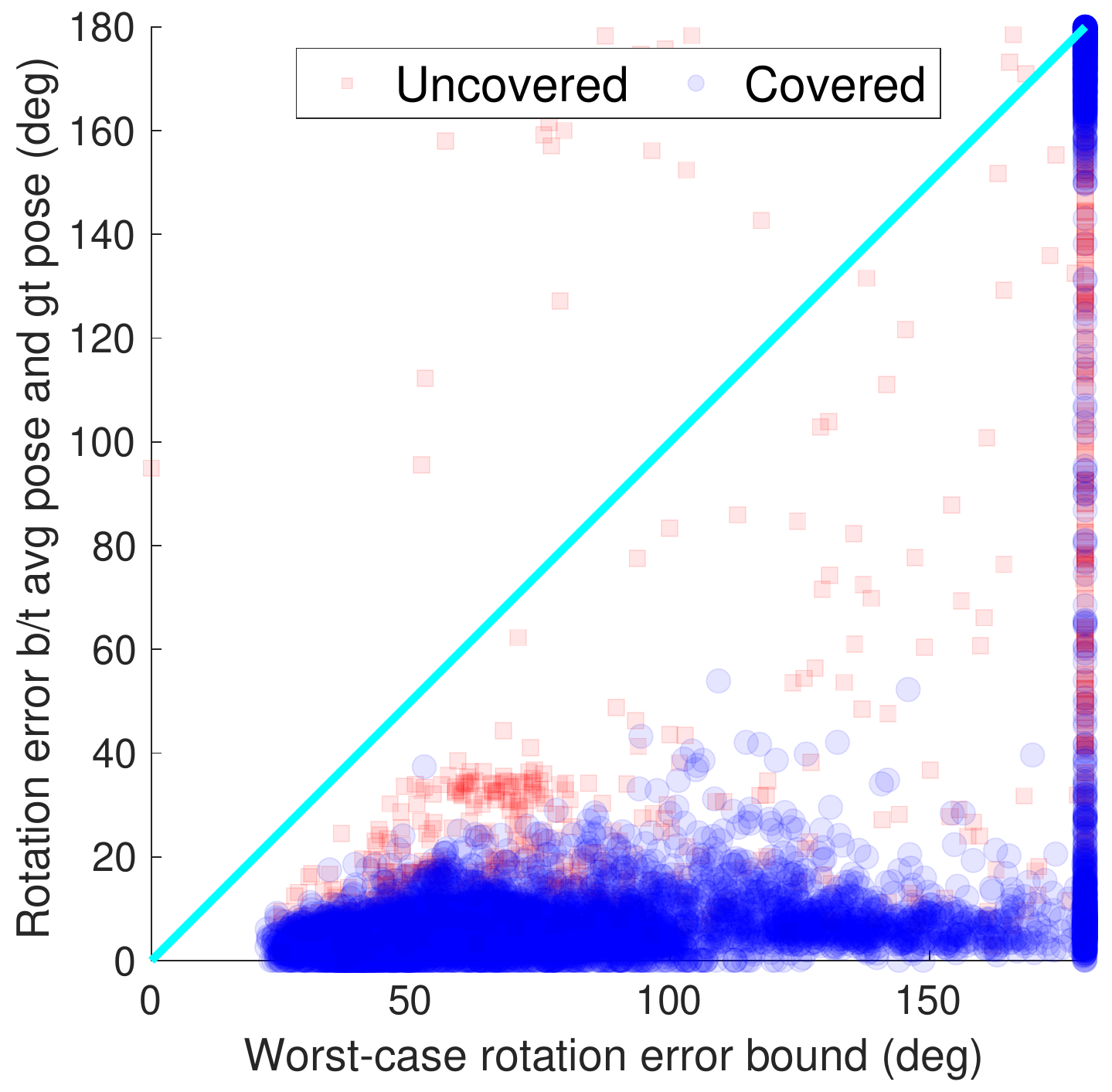}
		\end{minipage}
	&  	
	    \begin{minipage}{4cm}%
		\centering%
		\includegraphics[width=\columnwidth]{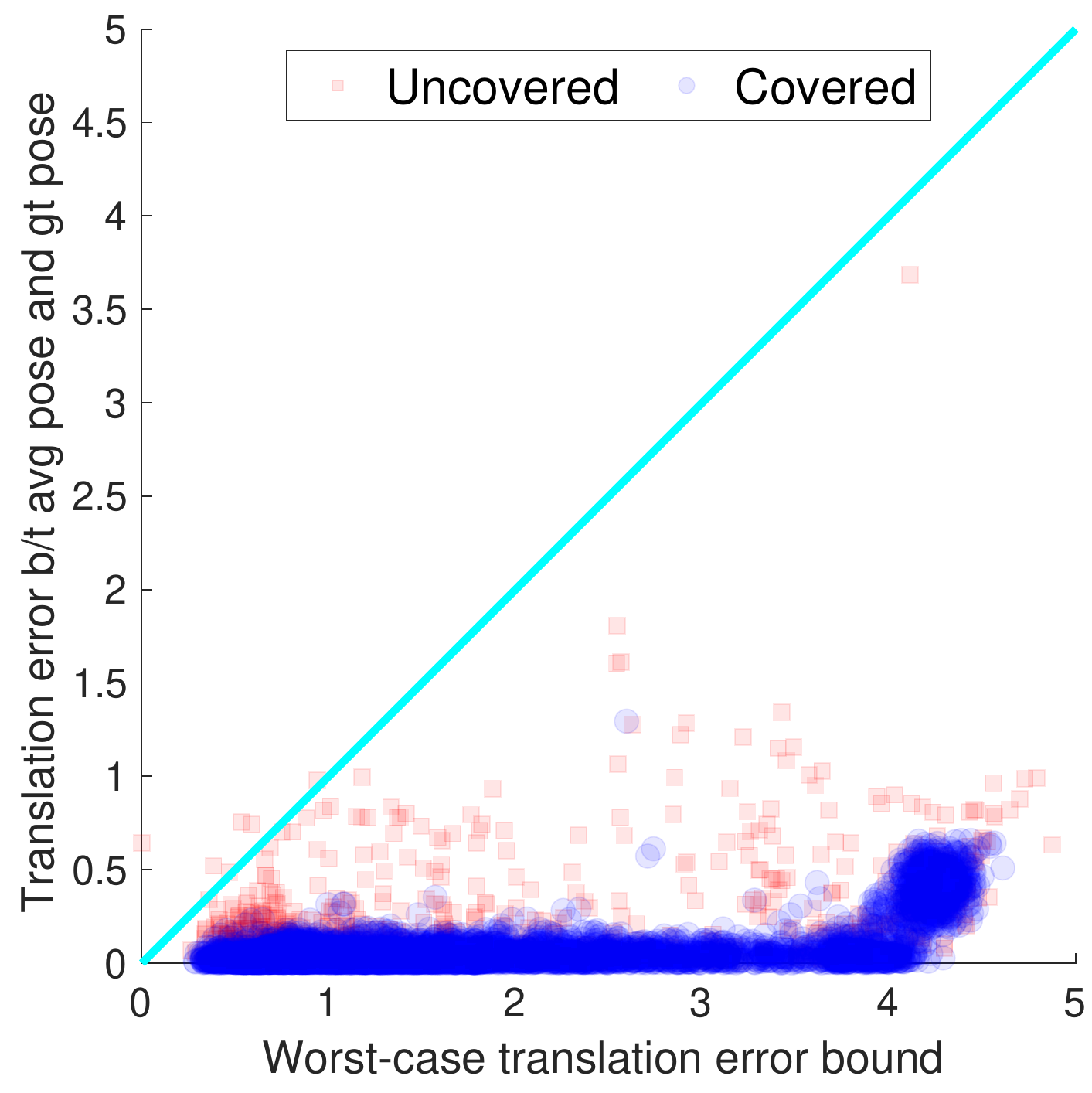}
		\end{minipage}
	&
	\begin{minipage}{4cm}%
		\centering%
		\includegraphics[width=\columnwidth]{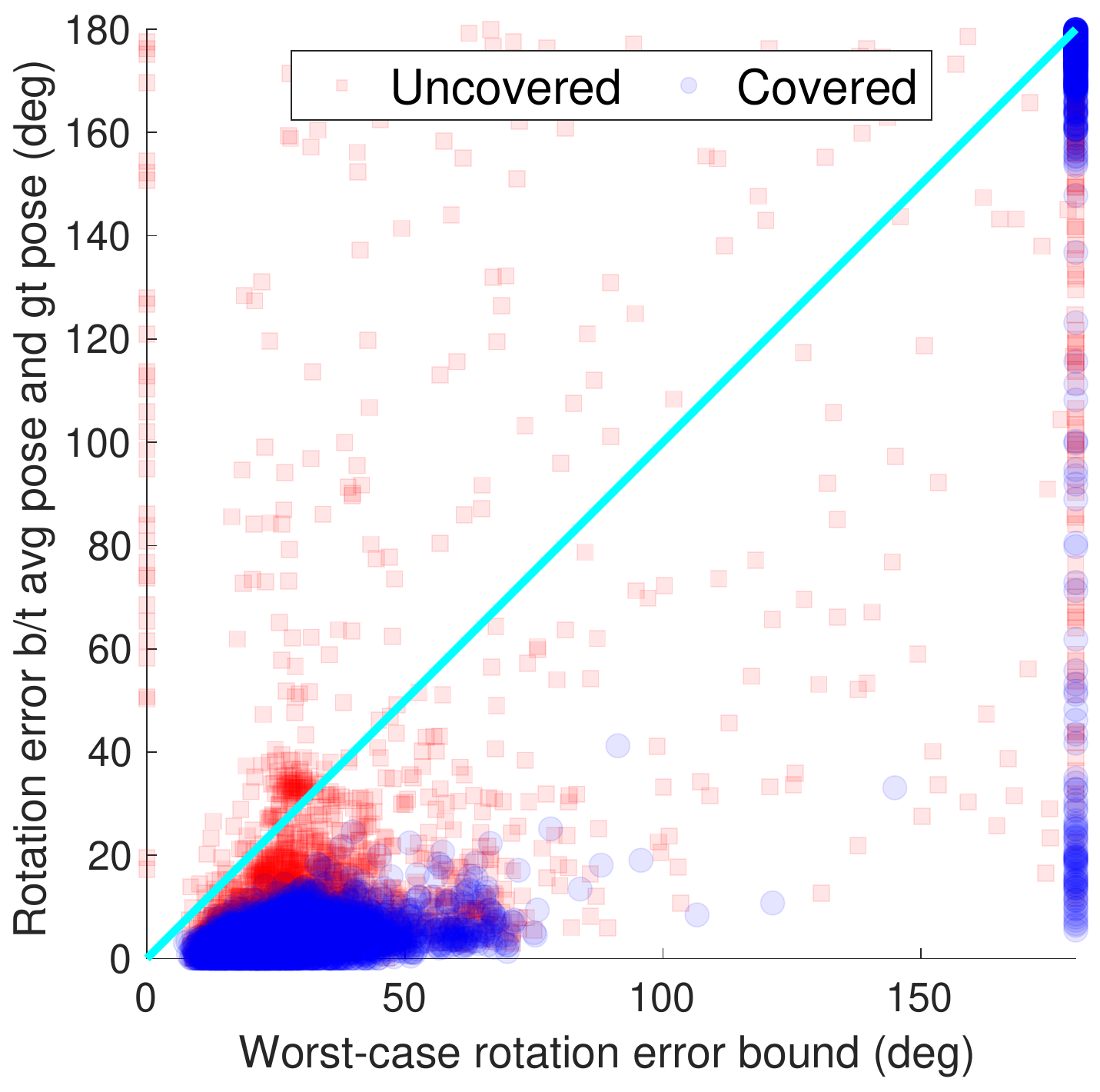}
		\end{minipage}
	&  	
	    \begin{minipage}{4cm}%
		\centering%
		\includegraphics[width=\columnwidth]{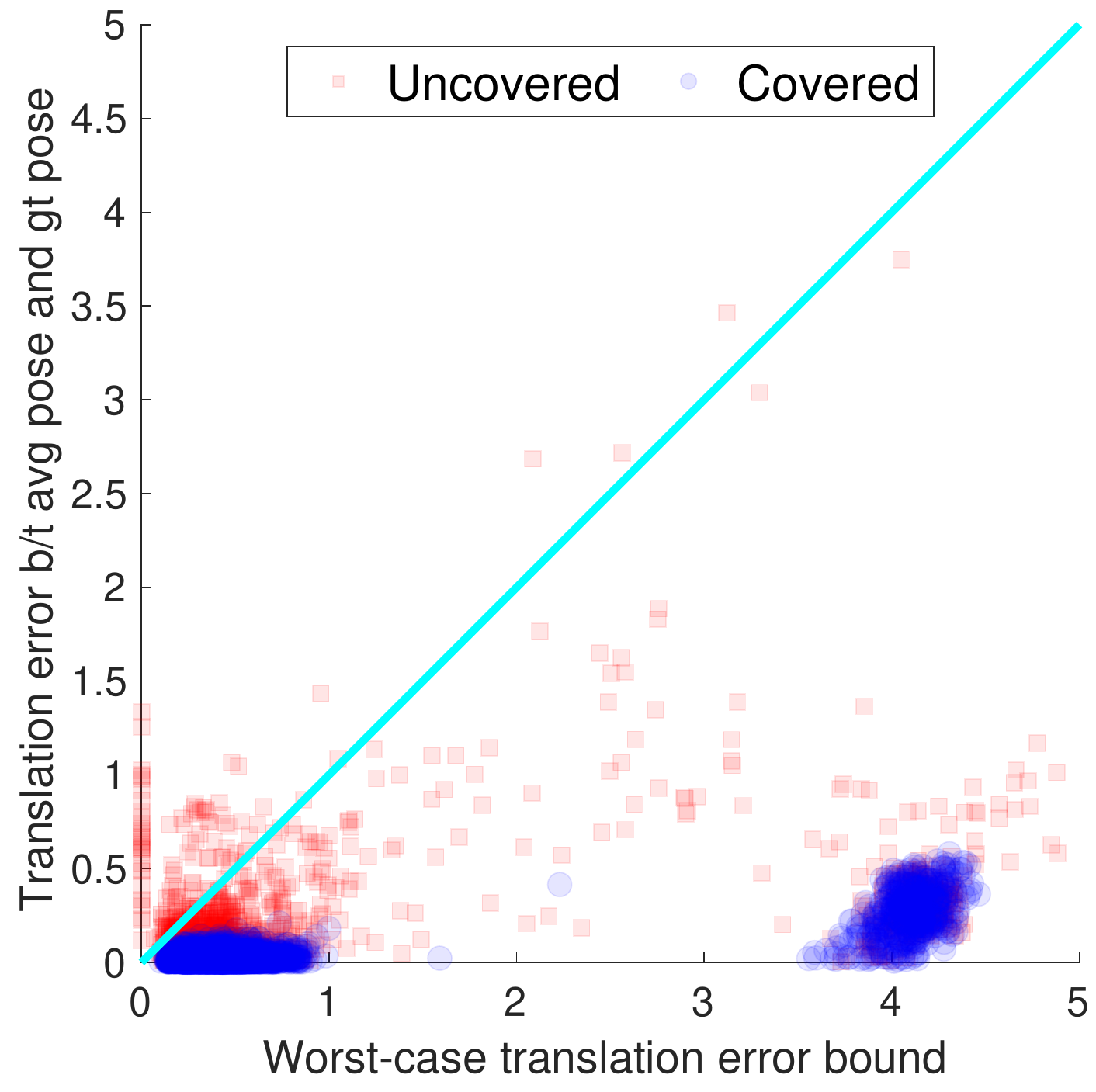}
		\end{minipage}
	\\
	\multicolumn{4}{c}{(c) \frcnnellipse. Left two columns: $\epsilon=0.1$; right two columns: $\epsilon=0.4$. }
\end{tabular}
\end{minipage}
\vspace{-4mm}
\caption{ Worst-case error bounds under (a) \gtellipse, (b) \frcnnball, and (c) \frcnnellipse setups. $x$-axis represents the worst-case error bounds computed from~\eqref{eq:pose2purse}, $y$-axis represents the actual error between average pose and groundtruth pose. The area below the diagonal $y=x$ indicates correctness of the bounds (\ie, bound $\geq$ error), and points that are closer to the diagonal from below indicate \emph{tighter} bounds (perfect if precisely lie on the diagonal). Blue circles plot cases where the {\purse} covers the groundtruth pose and red squares plot cases were the {\purse} does not cover the groundtruth. Notice that blue circles never cross the diagonal and our bounds are correct when the \purse contains the pose (which holds with $1-\epsilon$ marginal probability).\label{fig:other-error-bounds}} 
\end{center}
\vspace{-7mm}
\end{figure}

Fig.~\ref{fig:coverage-and-bound} middle and right columns (from the main document) show the worst-case error bounds (computed from~\eqref{eq:pose2purse}) of the average pose under the {\gtball} setup. Fig.~\ref{fig:other-error-bounds} shows the worst-case error bounds under the {\gtellipse}, {\frcnnball}, and {\frcnnellipse} setups, which are qualitatively similar to Fig.~\ref{fig:coverage-and-bound}. Notice that the blue circles never cross the $y=x$ diagonal, indicating our bounds are always valid when the \purse contains the groundtruth pose.

\subsection{A Closer Look at the Conservative Error Bounds}
The reader may have noticed two unusual results in the experiments on \lmo. First, the success rate on eggbox is consistently lower than other categories in our methods and other baselines (\eg, PVNet achieves $8.43\%$ success rate on eggbox, while the second lowest success rate is $55.37\%$). Second, the worst-case error bounds can be overly conservative, \eg, having $180^\circ$ rotation error bounds. It turns out both unusual results can be explained by the same reason: a labelling discrepancy in the {\lmo} dataset about eggbox.

We noticed the low success rate on eggbox across all baseline methods and contacted the authors of~\cite{schmeckpeper22jfr-semantic}, who encountered the same problem. One author told us ``\emph{I think this is a mistake or discrepancy in the 6DoF annotations of the dataset itself. As it [the eggbox] is considered a symmetric object, annotators for LMOD might not have consistently annotate it}''. Though it is possible to revise the nonconformity score for symmetric objects, the manually chosen keypoints by~\cite{schmeckpeper22jfr-semantic} break the symmetry. Therefore we decided to leave this discrepancy as is because it does not affect our probabilistic guarantees.

\begin{figure}
    \includegraphics*[width=0.49\linewidth]{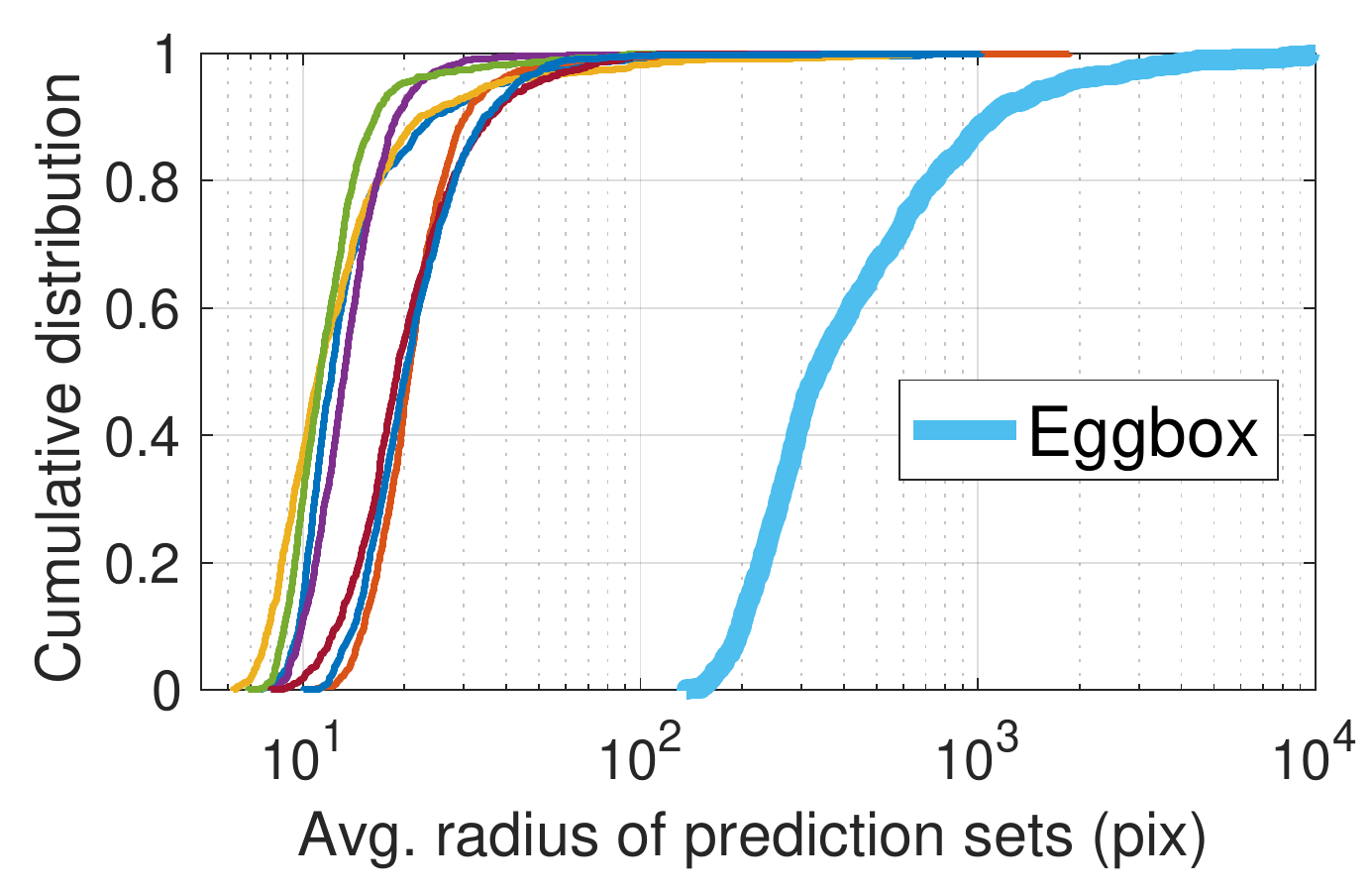}
    \hfill
    \includegraphics*[width=0.4\linewidth]{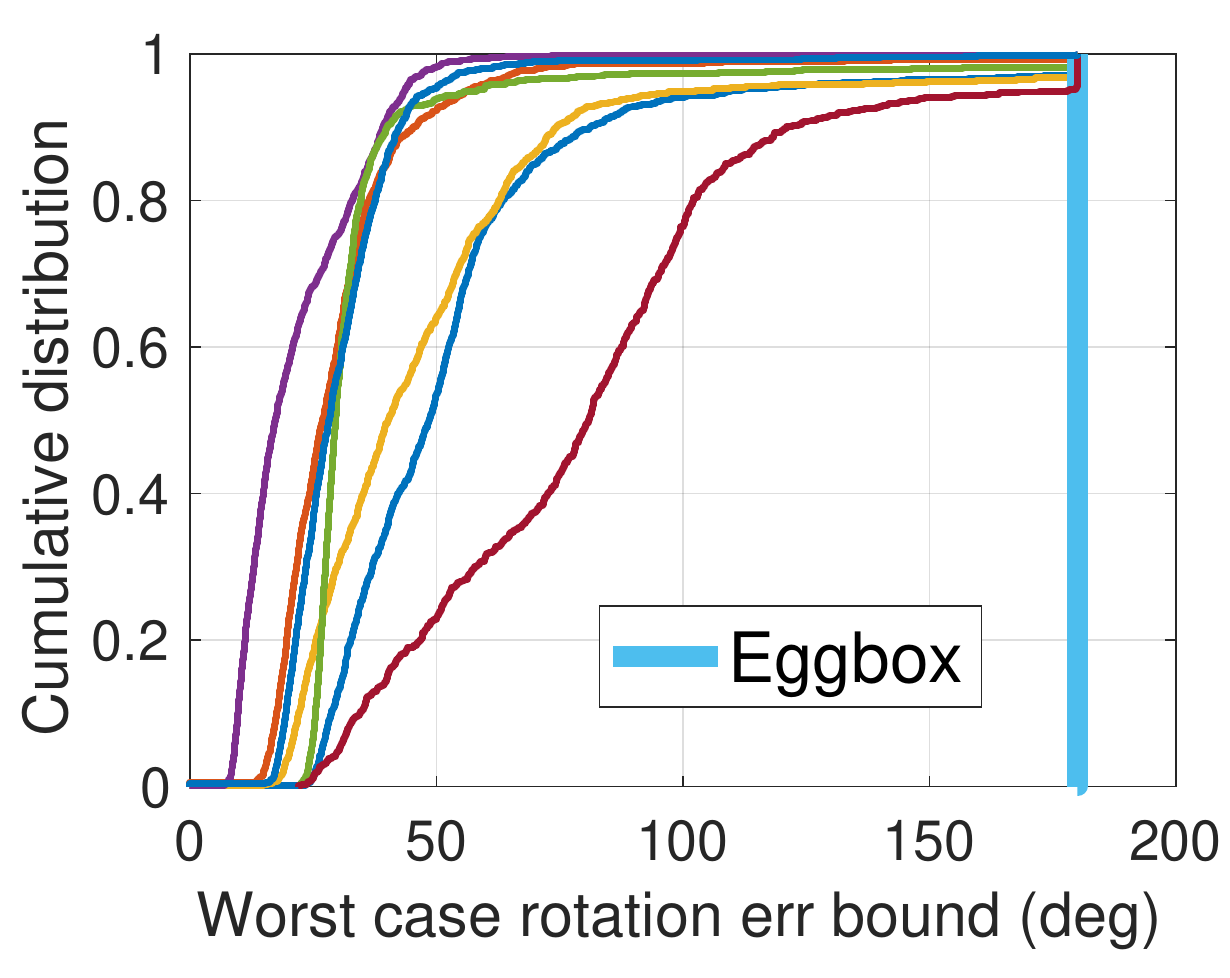}
    \vspace{-3mm}
    \caption{Left: cumulative distribution (CDF) of the average radius of prediction sets (under \gtball).
    Right: CDF of the worst-case rotation error bounds; \emph{eggbox} error bounds are mostly $180^\circ$. \label{supp:fig:eggbox}}
    \vspace{-4mm}
 \end{figure}

This labeling discrepancy, however, does translate to \emph{ conservative} prediction sets for the eggbox, in order to contain the (wrong) groundtruth at the desired probability. Fig.~\ref{supp:fig:eggbox} shows the eggbox prediction sets are \emph{one order of magnitude} larger than the other categories, leading to worst-case rotation error bounds being mostly $180^\circ$ (because the \purse is large enough to cover the entire $\SOthree$). {This indeed shows the advantage of our framework}: \emph{the user will see the large uncertainty produced by our algorithm and be alerted}!

Finally, because the \purse is too large, \ransag essentially 
returns a random sample in $\SOthree$, which has zero probability being close to the (wrong) groundtruth. Hence, a $0\%$ success rate makes sense. 

\subsection{Best Worst-case Error Bounds from Samples (Remark~\ref{rmk:bestbound})}

In Remark~\ref{rmk:bestbound}, we discussed that since solving~\eqref{eq:pose2purse} can provide worst-case error bounds for any pose estimator, the natural question is to ask if we can find better pose estimators (than the average pose computed from {\ransag}) with tighter worst-case error bounds, which boils down to solving the minimax problem in~\eqref{eq:bestbound}. However, problem~\eqref{eq:bestbound} is much more challenging to solve than~\eqref{eq:pose2purse}, and to the best of our knowledge, there is no efficient way to obtain a globally optimal solution. We think a good future research direction may be to explore methods in~\cite{pineda22neurips-theseus} or~\cite{nie17siopt-bilevel} for solving~\eqref{eq:bestbound}.

In this section, we provide a very preliminary study to explore if~\eqref{eq:bestbound} can indeed offer us tighter error bounds. Towards this goal, we randomly select $M=5$ pose samples $\{(R_i,t_i)\}_{i=1}^M$ from the results of {\ransag} (recall \ransag not only returns an average pose, but also returns a set of poses), and compute
\bea\label{eq:sampleworstbound}
\underline{d}^2_{\epsilon,\lambda} = \min \cbrace{  {d}^2_{i, \epsilon,\lambda} = \max_{(R,t) \in \Seps} \lambda \Fnorm{R - R_i}^2 + (1-\lambda) \norm{ t - t_i}^2  }_{i=1}^M,
\eea
which first solves~\eqref{eq:pose2purse} (inner ``$\max$'' in~\eqref{eq:sampleworstbound}) for each $(R_i,t_i)$ and then selects the minimum (tightest) error bounds. Note that we still apply a second-order SDP relaxation when computing the error bounds for each $(R_i,t_i)$ since~\eqref{eq:pose2purse} is nonconvex. 

\begin{figure}[h]
\begin{center}
\begin{minipage}{\textwidth}
\centering
\begin{tabular}{cc}%
	    \begin{minipage}{5cm}%
		\centering%
		\includegraphics[width=\columnwidth]{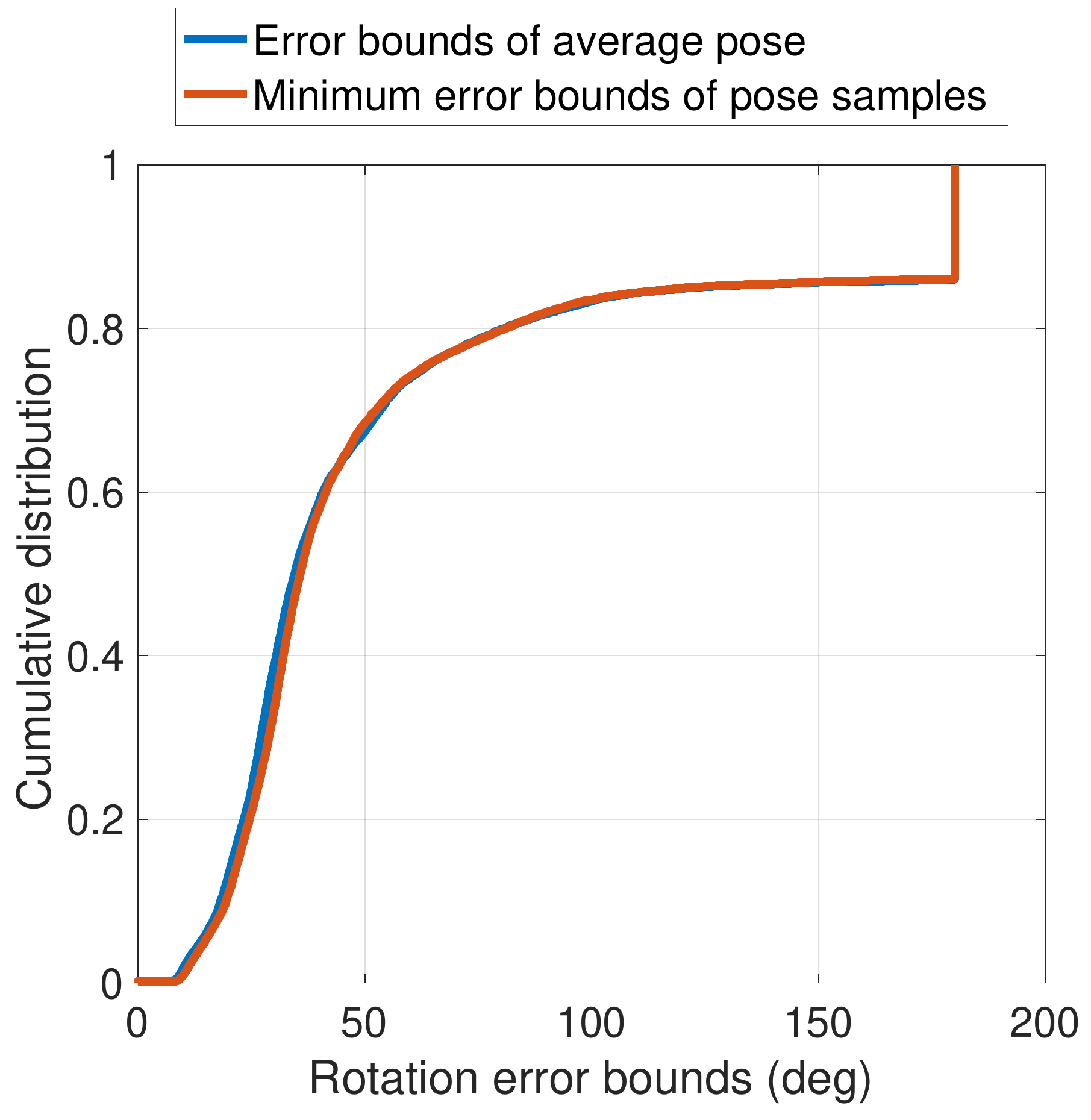}
		\end{minipage}
	&  	
	    \begin{minipage}{5cm}%
		\centering%
		\includegraphics[width=\columnwidth]{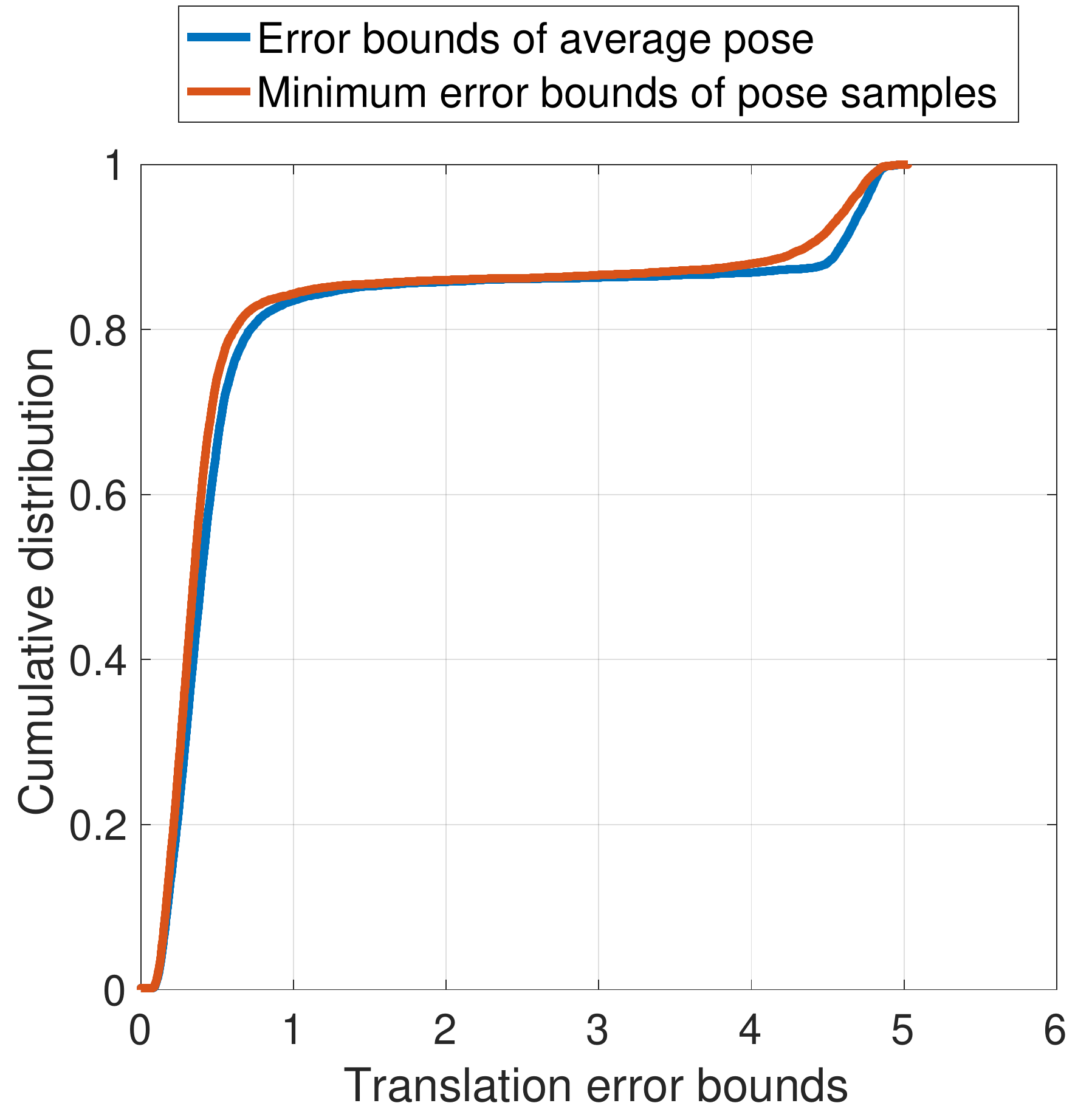}
		\end{minipage}
\end{tabular}
\end{minipage}
\vspace{-4mm}
\caption{Cumulative distribution function (CDF) of the worst-case error bounds under the {\gtball} setup with $\epsilon=0.1$. Blue curve plots the CDF of the error bounds of the average pose, and red curve plots the CDF of the minimum error bounds of the pose samples (\ie solving~\eqref{eq:sampleworstbound}). We can see that the translation error bounds are slightly tightened by selecting the minimum error bounds for multiple pose samples. \label{fig:avg-vs-smp}} 
\end{center}
\vspace{-7mm}
\end{figure}

Fig.~\ref{fig:avg-vs-smp} plots the cumulative distribution functions (CDF) of the error bounds under the {\gtball} setup with $\epsilon=0.1$. The blue curves plot the CDF of the error bounds computed for the average pose, while the red curves plot the CDF of the error bounds computed from solving~\eqref{eq:sampleworstbound}. We can see that solving~\eqref{eq:sampleworstbound} does slightly improve the tightness of the translation bounds (while the rotation bounds are very close). Considering that we only select the minimum error bounds from $M=5$ samples, we conjecture solving the minimax problem can give us much tighter error bounds, and we leave this as an exciting future research.

{\small
\bibliographystyle{ieee_fullname}
\bibliography{refs}
}

\end{document}